\documentclass[11pt]{article}

\usepackage[preprint]{acl}
\usepackage{times}
\usepackage{latexsym}
\usepackage{xspace}
\usepackage[T1]{fontenc}
\usepackage[utf8]{inputenc}
\usepackage{microtype}
\usepackage{inconsolata}
\usepackage{graphicx}
\usepackage{paralist}
\usepackage{amssymb}
\usepackage{amsmath}
\usepackage{booktabs}
\usepackage{multirow}
\usepackage{hyperref}
\usepackage{xurl}

\AtEndPreamble{
    \usepackage[capitalize]{cleveref}
    \crefname{section}{Sec.}{Secs.}
    \Crefname{section}{Section}{Sections}
    \Crefname{table}{Table}{Tables}
    \crefname{table}{Tab.}{Tabs.}
}

\makeatletter
\DeclareRobustCommand\onedot{\futurelet\@let@token\@onedot}
\def\@onedot{\ifx\@let@token.\else.\null\fi\xspace}
\def\eg{\emph{e.g}\onedot}

\newcommand{\mymethod}{\texttt{Syn-Omni}}
\newcommand{\baseline}{\texttt{Uni-Omni}}

\usepackage{tikz}
\usetikzlibrary{positioning, fit, backgrounds, arrows.meta, shapes.geometric, calc}

\title{
Syn-Omni: Structured Specialization and 

Progressive Collaboration for Omnimodal Embeddings
}

\newcommand{\mailto}[2]{\href{mailto:#1}{\color{black}#2}}
\author{%
    Youngtaek Oh\textsuperscript{1,2}\thanks{Work done during internship at Sony Group Corporation}\quad%
    Qiyu Wu\textsuperscript{2}\thanks{Corresponding Author:~\texttt{qiyu.wu@sony.com}}\quad%
    Hiromi Wakaki\textsuperscript{2}\quad%
    Junmo Kim\textsuperscript{1}\quad%
    Yuki Mitsufuji\textsuperscript{2}\\
    \textsuperscript{1}KAIST \qquad%
    \textsuperscript{2}Sony Group Corporation \\
    \textsuperscript{1}\texttt{%
            \{%
                \mailto{youngtaek.oh@kaist.ac.kr}{youngtaek.oh}, %
                \mailto{junmo.kim@kaist.ac.kr}{junmo.kim}%
            \}@kaist.ac.kr%
    }\\%
    \textsuperscript{2}\texttt{%
            \{%
                \mailto{qiyu.wu@sony.com}{qiyu.wu}, %
                \mailto{hiromi.wakaki@sony.com}{hiromi.wakaki}, %
                \mailto{yuhki.mitsufuji@sony.com}{yuhki.mitsufuji}%
            \}@sony.com%
    }\\%
    \url{https://github.com/sony/syn-omni}
}%
\begin{document}
\maketitle
\begin{abstract}
Omnimodal embeddings naturally involve both shared representations and modality-specific features across heterogeneous inputs. However, existing omnimodal embedding methods often rely on a single shared parameter space over mixed-modality data, limiting structural separation between universal and modality-specific representations. To address this, we propose \mymethod, a unified framework for structured omnimodal adaptation with modality specialization and controlled cross-modal collaboration. Specifically, we introduce Orthogonal Modality-Expert LoRA (OME-LoRA), which decomposes adaptation into a shared LoRA path for universal semantics and modality-expert LoRA paths for modality-aware specialization. Furthermore, Progressive Synergy Routing (PSR) enables experts to first establish modality-specific priors, then gradually interact with other modality-experts for cross-modal synergy. Evaluated across 81 diverse tasks spanning image, video, audio, and audiovisual modalities, \mymethod\ consistently outperforms omnimodal baselines, demonstrating the effectiveness of structured specialization and cross-modal progressive collaboration.
\end{abstract}

\addtocontents{toc}{\protect\setcounter{tocdepth}{-1}}

\section{Introduction}
Omnimodal Large Language Models (Omni-MLLMs) have opened a new frontier toward unified understanding across modalities, including text, images, video, audio, and their composed inputs~\cite{xu2025qwen25}. With these versatile foundations, recent studies~\cite{xiao2026scaling,chen-etal-2026-e5} have successfully repurposed Omni-MLLMs into discriminative embedding models through contrastive fine-tuning~\cite{jiang2025vlmvec,meng2026vlmvecv} with LoRA~\cite{hu2022lora}, offering an efficient and scalable framework for unified embedding spaces across heterogeneous modalities.

\begin{figure}[t]
  \centering
   \includegraphics[width=1.0\columnwidth]{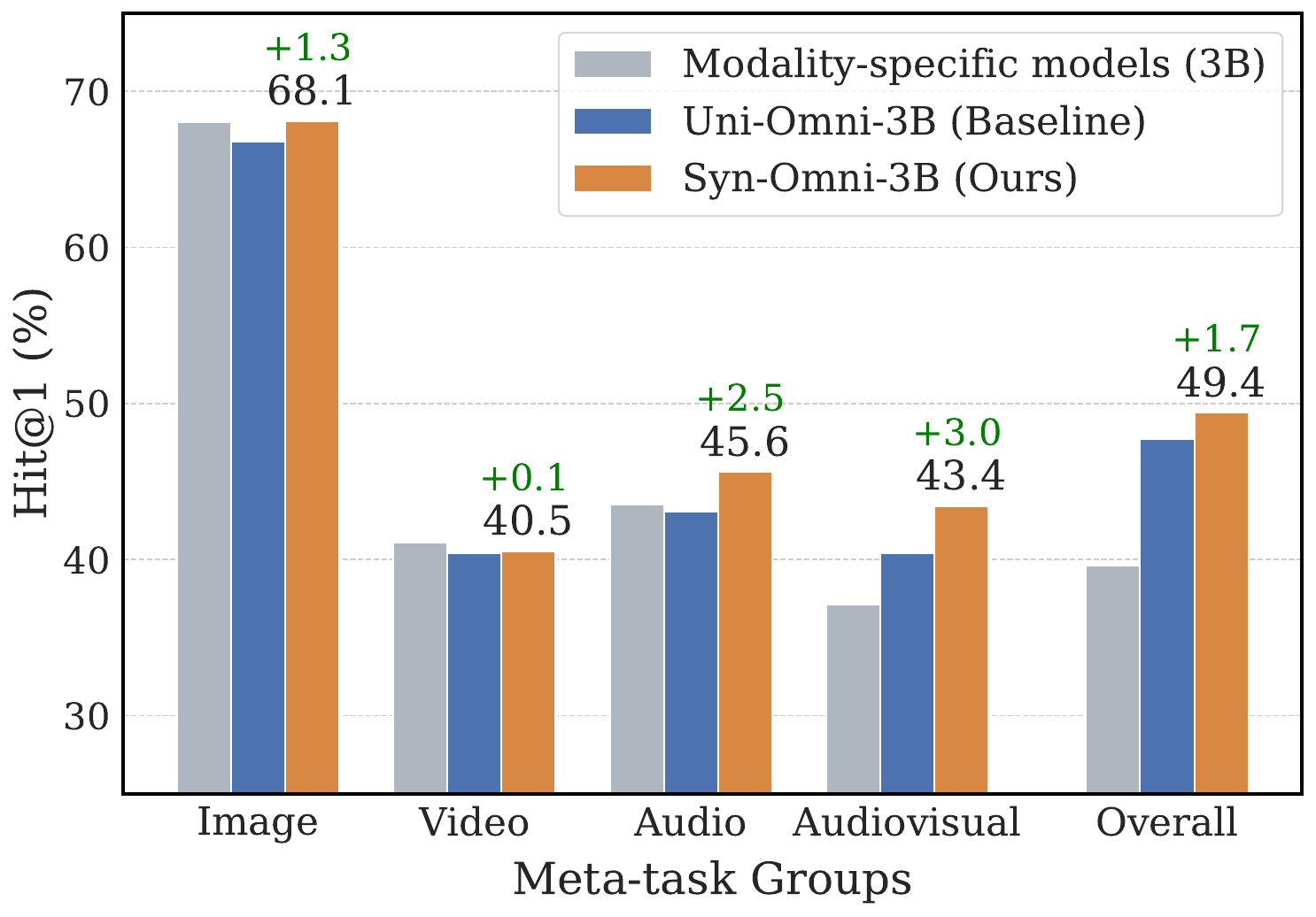}
    \caption{\textbf{Omnimodal adaptation strategies.} 
    We compare three contrastive embedding setups: 
    \textbf{(1)} modality-specific models, each trained independently on a single modality using a shared LoRA, \textbf{(2)} \baseline\ (baseline), trained on mixed-modality data using a single shared LoRA, and \textbf{(3)} \mymethod\ (Ours), which combines a shared adaptation path with modality-specialized expert paths. Here, the overall score of the modality-specific setup averages each model across all four modalities, including those it was not trained on. While mixed-modality training enables strong unified representations, \mymethod\ further improves the \baseline\ baseline across diverse modality scenarios. 
    }
   \label{fig:teaser}
\end{figure}

Despite their strong performance, existing omnimodal embedding methods largely rely on fully shared parameter adaptation over mixed-modality data. While such shared adaptation is effective in learning a unified representation space, it provides limited structural separation between modality-shared and modality-specific representations. Since different modalities exhibit both common semantic structures and distinct modality characteristics, a shared adaptation space may underutilize modality-aware specialization over the course of training.

As shown in \cref{fig:teaser}, different omnimodal adaptation strategies exhibit complementary strengths. Modality-specific models effectively preserve modality-aware characteristics, but their isolated training restricts cross-modal interaction and limits generalization. In contrast, \baseline, a baseline trained on mixed-modality data with a single shared LoRA, learns a unified embedding space that performs well on composed multimodal settings such as audiovisual tasks. However, such fully shared adaptation provides limited structural separation to model modality-specific characteristics, whereas completely isolated experts fail to leverage cross-modal synergy. These observations highlight the need for a structured omnimodal adaptation strategy that can jointly support a unified semantic foundation and modality-aware specialization while encouraging collaborative interactions within an omnimodal embedding space.

To this end, we propose \mymethod, a framework that structurally separates shared and modality-specialized expert paths while progressively enabling cross-modal collaboration. Our framework consists of two key components: \textit{Orthogonal Modality-Expert LoRA} (OME-LoRA) and \textit{Progressive Synergy Routing} (PSR). OME-LoRA separates adaptation into shared and modality-specific LoRA paths to capture both universal semantics and distinct characteristics. Complementing this, PSR gradually enables these isolated modality-experts to interact and establish cross-modal synergy during training. As shown in \cref{fig:teaser}, \mymethod\ consistently improves upon the unified baseline, showing improved gains on composed multimodal tasks such as audiovisual understanding.

We extensively evaluate \mymethod\ across 81 diverse embedding tasks spanning image, video, audio, and audiovisual modalities, covering a broad range of task types including classification, retrieval, and QA. We show that \mymethod\ outperforms existing omnimodal embedding baselines while achieving more balanced performance across these diverse modality settings. These results demonstrate the effectiveness of structured specialization and progressive cross-modal collaboration for unified omnimodal representation learning.

We summarize our contributions as follows:
\vspace{-2mm}
{\setdefaultleftmargin{1mm}{}{}{}{}{}
\begin{itemize}
  \item We address the structural limitations of existing fully shared or entirely isolated omnimodal adaptation methods by presenting a unified framework that simultaneously balances robust semantic foundations and modality-aware specialization.
  \vspace{-2mm}
  
  \item We propose \mymethod, which pairs structured shared-expert adaptation in OME-LoRA with progressive cross-modal collaboration in PSR, where experts first establish modality priors and then gradually interact with other modalities.
  \vspace{-2mm}
  
  \item \mymethod\ is evaluated across 81 tasks in image, video, audio, and audiovisual modalities, with various task types. Results show that \mymethod\ improves baselines across diverse modalities.
\end{itemize}
}

\section{Related Work}

\textbf{Omnimodal Embeddings.}
Multimodal Large Language Models (MLLMs) have been repurposed into embedding models through contrastive fine-tuning. Early frameworks on vision-language models like E5-V~\cite{jiang2024e5} and VLM2Vec~\cite{jiang2025vlmvec} pioneered this transition by applying LoRA~\cite{hu2022lora} to text-image retrieval. Covering broader modalities, GVE~\cite{guo2025towards} and VLM2Vec-V2~\cite{meng2026vlmvecv} extended this paradigm to video, while AuroLA~\cite{xu2026scaling} focused on audio-text retrieval. Furthermore, Omni-Embed-Nemotron~\cite{xu2025omni} and WAVE~\cite{tang2026wave} enable any-to-any retrieval across text, image, audio, and video by leveraging large omnimodal backbones~\cite{xu2025qwen25}. To further refine cross-modal alignment, methods such as LCO-Emb~\cite{xiao2026scaling} and e5-omni~\cite{chen-etal-2026-e5} introduce explicit calibration strategies, including language-centric refinement and modality-aware temperature scaling. While most existing approaches rely on fully-shared adaptation (\eg, a single LoRA layer) across modalities, we introduce a decoupled adaptation framework in the omnimodal setting.

\begin{figure*}[t]
  \centering
   \includegraphics[width=0.97\linewidth]{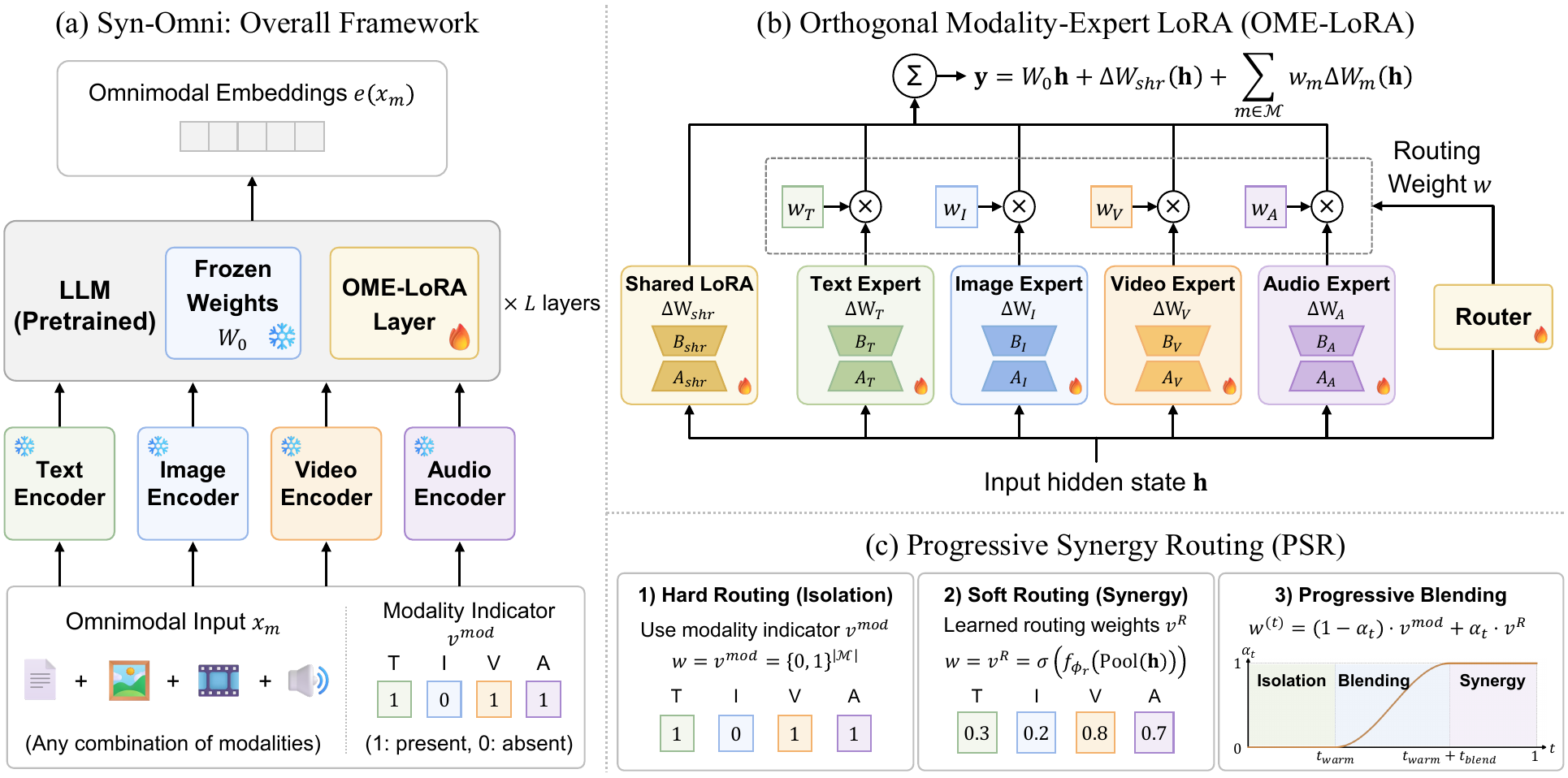}
   \caption{
   \textbf{Illustration of the \mymethod\ framework.} 
\textbf{(a)} Overall Framework: Omnimodal inputs $x_m$ are encoded by modality-specific encoders and a pretrained LLM backbone with OME-LoRA to produce unified embeddings $e(x_m)$.
\textbf{(b)} Orthogonal Modality-Expert LoRA (OME-LoRA): Shared LoRA path captures universal semantics, while modality-specific expert LoRAs model modality-aware characteristics. The outputs of modality-experts are combined by the routing weight $w$.
\textbf{(c)} Progressive Synergy Routing (PSR): PSR gradually shifts from initial modality-separate routing to later-stage collaborative routing, enabling progressive cross-modal synergy.}
   \label{fig:overall}
\end{figure*}

\noindent \textbf{Structured Adaptation.}
Several approaches explicitly separate parameters to encourage specialization. MoELoRA~\cite{luo2024moelora} and MoDULA~\cite{ma-etal-2024-modula} instantiate task-level experts within LoRA, while a separate line imposes orthogonality on the internal subspaces of each adapter~\cite{zhang2023adaptive,li2025stella}. Beyond architectural separation, LIMoE~\cite{mustafa2022multimodal} shows that emergent modality-specific experts require explicit control for stable training. DGL~\cite{wei2025boosting} shows that aggressive fusion can suppress modality-specific learning, while MCA~\cite{wu2025mca} reports shortcut behaviors induced by simple contrastive objectives. From an optimization perspective, Uni-X~\cite{hao2026unix} reveals the limitations of shared transformers across heterogeneous modalities. We adapt these ideas to omnimodal embedding learning, separating experts along the modality axis rather than the task axis and regularizing the shared-expert relation instead of each adapter's internal subspace.

\section{Syn-Omni Framework}
We propose \mymethod, a unified omnimodal embedding framework that moves beyond fully shared adaptation across modalities. As shown in~\cref{fig:overall}, our approach consists of two main components that jointly realize modality specialization and progressive cross-modal collaboration: \textit{Orthogonal Modality-Expert LoRA} (OME-LoRA) and \textit{Progressive Synergy Routing} (PSR). 
We first introduce our omnimodal embedding formulation \baseline\ in~\cref{sec:3_1}.
We then detail the design of OME-LoRA in \cref{sec:3_2}, the PSR mechanism in \cref{sec:3_3}, and the training objectives in \cref{sec:3_4}.

\subsection{Omnimodal Embedding Learning}
\label{sec:3_1}
We consider a mixed omnimodal training setting, where we construct a heterogeneous training mixture encompassing various modality combinations. Under this setup, we train \baseline, a simple contrastive baseline that adapts a pretrained Omni-MLLM using a single modality-shared LoRA. 

\noindent \textbf{Omnimodal Representation.}
We define a base modality set $\mathcal{M} = \{T, I, V, A\}$, representing text, image, video, and audio, respectively. An input $x_m$ is any non-empty subset $m \subseteq \mathcal{M}$, and we define a modality indicator $v^{mod} \in \{0, 1\}^{|\mathcal{M}|}$ indicating the presence of each modality in $x_m$.
To obtain the omnimodal embedding $\mathbf{e}(x_m)$ from a pretrained Omni-MLLM $f_\theta$, the input is processed with an instruction prompt, and last-token pooling is applied from the final layer:
$\mathbf{e}(x_m) = \text{Pool}(f_\theta(x_m))$.

\noindent \textbf{Contrastive Optimization via LoRA.}
The pretrained weights $W_0 \in \theta$ are frozen, while low-rank adapters $\phi$ are trained. For an intermediate hidden state $\mathbf{h}$, the adapted output is
$\mathbf{y} = W_0 \mathbf{h} + \Delta W_{\phi}(\mathbf{h})$.
The model is optimized with an InfoNCE objective to align matched pairs in the unified embedding space. For a batch of $N$ pairs, the loss is:
\begin{equation}
\mathcal{L}_{con} = -\frac{1}{N} \sum_{i=1}^{N} \log \frac{\exp(\text{sim}(\mathbf{e}_i, \mathbf{e}_i^+) / \tau)}{\sum_{j=1}^{N} \exp(\text{sim}(\mathbf{e}_i, \mathbf{e}_j) / \tau)},
\end{equation}
where $\text{sim}(\cdot, \cdot)$ denotes cosine similarity and $\tau$ is a temperature hyperparameter. We further detail our omnimodal training mixture in~\cref{sec:train_data}.

\subsection{Orthogonal Modality-Expert LoRA}
\label{sec:3_2}
The training of omnimodal embeddings involves heterogeneous modalities with different characteristics, yet typically relies on a single shared LoRA applied across all modalities. This motivates a more structured design to better organize shared and modality-specific adaptation. To this end, we propose \textit{Orthogonal Modality-Expert LoRA} (OME-LoRA), which introduces a structured decomposition of shared and modality-expert LoRA paths.

\noindent \textbf{Parallel Shared-Expert LoRA.}
Inspired by the decoupling of universal and domain-specific knowledge in multi-task learning literature~\citep{ma-etal-2024-modula}, we extend this foundation to the omnimodal setting to jointly model shared semantics and modality-specialized characteristics. Specifically, as shown in~\cref{fig:overall} (b), a shared LoRA path is consistently activated for all input modalities, while modality-expert LoRA paths are selectively engaged based on the modalities present in $x_m$. For an intermediate hidden state $\mathbf{h}$, the output of the OME-LoRA layer $\mathbf{y}$ is computed as follows:
\begin{equation}
\label{eqn:ome_lora}
\mathbf{y} = W_0 \mathbf{h} + \Delta W_{shr}(\mathbf{h}) + \sum_{m \in \mathcal{M}} w_m \Delta W_m(\mathbf{h}),
\end{equation}
where $\Delta W_{shr}$ captures modality-shared parameter updates and $\Delta W_m$ models modality-specific counterparts. The routing weights $w_m$ enable adaptive integration of modality-expert outputs. We defer the details of modality-aware specialization in modality-experts and the routing strategy for cross-modal synergy among experts to~\cref{sec:3_3}.

\noindent \textbf{Decoupling via Orthogonal Penalty.}
We introduce an orthogonal penalty $\mathcal{L}_{ortho}$ that encourages modality-specific experts to learn complementary weight directions relative to the shared pathway. Given that each LoRA update is decomposed into low-rank matrices $\Delta W = BA$, we define the vectorized form of the input-side projection matrix as $\mathbf{a} = \text{vec}(A)$.
The penalty is then defined as:
\begin{equation}
\mathcal{L}_{ortho} = \frac{1}{|\mathcal{M}|} \sum_{m \in \mathcal{M}} \left( \frac{|\tilde{\mathbf{a}}_{shr} \cdot \mathbf{a}_m|}{\|\tilde{\mathbf{a}}_{shr}\| \|\mathbf{a}_m\|} \right)^2,
\end{equation}
where $\mathcal{L}_{ortho}$ minimizes the cosine similarity between the weight directions of the shared and expert LoRA parameters. Here, we apply the penalty only to $A$, which determines the input subspace of each path, leaving the output projection $B$ unconstrained. We also treat the shared component as a fixed reference by applying a stop-gradient operator, $\tilde{\mathbf{a}}_{shr} = \mathrm{sg}[\mathbf{a}_{shr}]$. This blocks gradient propagation so that the shared path is not affected by the penalty. In this way, the penalty helps reduce redundancy between the shared and expert paths.

\subsection{Progressive Synergy Routing}
\label{sec:3_3}
The interaction between modality-experts is governed by the routing weights $w_m$ in~\cref{eqn:ome_lora}. While hard routing based on the modality indicator $v^{mod}$ activates only the corresponding experts, it restricts cross-modal knowledge exchange across experts. In contrast, soft routing with a learnable router enables synergy, but it can blunt expert specialization if the experts lack sufficient modality-aware priors early on, as explored in~\cref{fig:similarity} (left).

To facilitate stable collaboration among experts, we propose \textit{Progressive Synergy Routing} (PSR), allowing experts to first establish modality priors via hard routing and then progressively exchange complementary information across modalities with soft routing, as illustrated in~\cref{fig:overall} (c). In addition, the Synergy Margin Loss (SML) is designed to avoid excessive suppression of absent-modality experts, preserving expert interactions necessary for cross-modal synergy.

\noindent \textbf{Progressive Blending Mechanism.}
We first introduce a learnable router $f_{\phi_r}$ that generates soft routing weights $\mathbf{v}^R = \sigma(f_{\phi_r}(\text{Pool}(\mathbf{h})))$ from the intermediate hidden state $\mathbf{h}$, where $f_{\phi_r}$ is a fully-connected layer with parameters $\phi_r$ and $\sigma(\cdot)$ denotes the sigmoid function. This soft router enables experts from non-input modalities to contribute dynamically to the final outputs. To resolve the trade-off between expert specialization and synergy, we propose to progressively blend the hard and soft routing weights over the training progress $t$ to smoothly shift from \textit{isolation} to \textit{interaction}. The blended routing weight $w_m^{(t)}$ is computed as:
\begin{equation}
w_m^{(t)} = (1 - \alpha_t) \cdot v_m^{mod} + \alpha_t \cdot v_m^R,
\end{equation}
where $\alpha_t \in [0, 1]$ is scheduled along with the training progress ratio $t$ (\eg, current step / total steps):
\begin{equation}
\alpha_t = 
\begin{cases} 
0, & 0 \le t < t_{\text{warm}} \\
g(t), & t_{\text{warm}} \le t < t_{\text{warm}} + t_{\text{blend}} \\
1, & t_{\text{warm}} + t_{\text{blend}} \le t \le 1.
\end{cases}
\end{equation}
Here, $g(t) = \frac{1}{2} [ 1 - \cos ( \pi \frac{t - t_{\text{warm}}}{t_{\text{blend}}} ) ]$ denotes a cosine-based monotonic transition function from 0 to 1, where $t_{\text{warm}}$ controls how long the experts remain isolated and $t_{\text{blend}}$ sets the length of the transition. The schedule consists of three phases: \textit{Warmup} ($t < t_{\text{warm}}$) for strict isolation with expert specialization, \textit{Blending} ($t_{\text{warm}} \leq t<t_{\text{warm}} + t_{\text{blend}}$) for a smooth transition between hard and soft routing, and \textit{Synergy} ($t \ge t_{\text{warm}} + t_{\text{blend}}$) for full dynamic collaboration via soft routing weights $\mathbf{v}^R$. This schedule can be viewed as a curriculum~\cite{bengio2009curriculum} over routing that eases optimization, where experts are first exposed to their own modality before interacting with others. To empirically capture this transition, \cref{fig:progress_routing} visualizes the evolution of routing weights over the course of training. 

\begin{figure}[t]
  \centering
   \includegraphics[width=0.92\columnwidth]{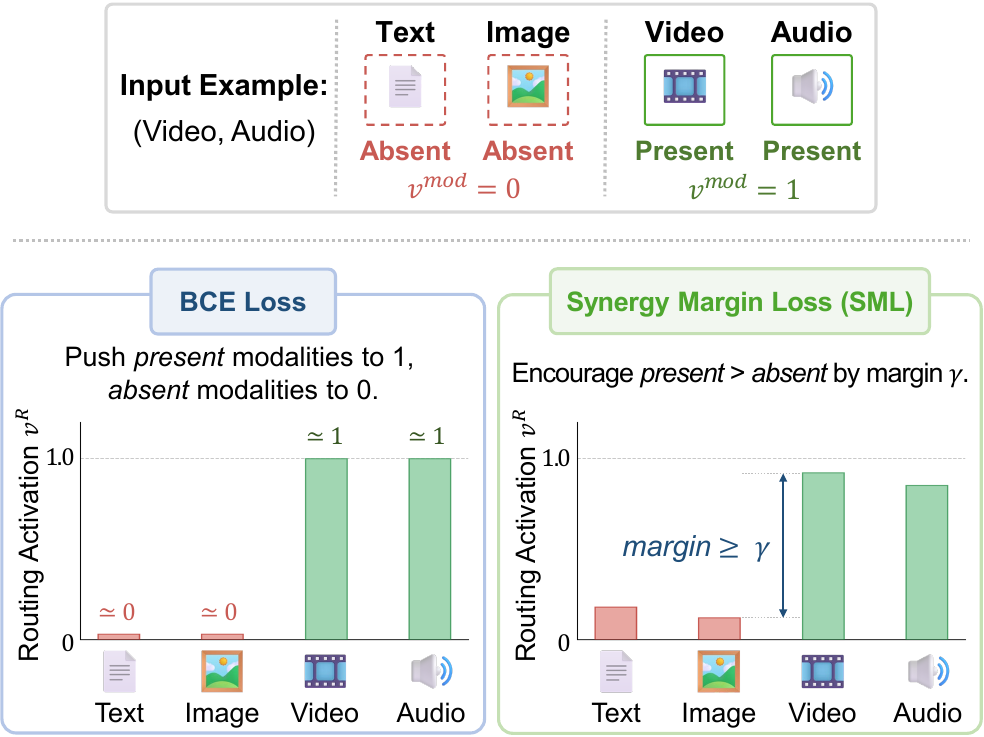}
    \caption{Conceptual comparison between BCE Loss and Synergy Margin Loss (SML). While BCE strictly suppresses absent modalities to 0, SML maintains a relative margin $\gamma$ ($v^R_{\text{present}} - v^R_{\text{absent}} \ge \gamma$), which allows absent-modality experts to retain meaningful activations, preserving cross-modal synergy during training.}
   \label{fig:sml_loss}
    \end{figure}

\noindent \textbf{Synergy Margin Loss.}
To encourage modality-aware specialization in the router, we introduce an auxiliary supervision that guides the soft routing activations $\mathbf{v}^R$ using the modality indicators ${v}^{mod}$. While Binary Cross-Entropy (BCE) could directly enforce modality-aligned routing, it tends to excessively suppress absent modalities, eliminating potentially useful interactions across experts. 
To preserve such collaborative capacity while maintaining the dominance of input-relevant modalities, we propose a Synergy Margin Loss $\mathcal{L}_{SML}$. Let $\mathcal{P}$ and $\mathcal{A}$ denote the sets of present ($v_m^{mod}=1$) and absent ($v_m^{mod}=0$) modality indices in the input:
\begin{equation}
\label{eqn:margin_synergy}
\mathcal{L}_{SML} = \frac{1}{|\mathcal{P}||\mathcal{A}|} \sum_{i \in \mathcal{P}} \sum_{j \in \mathcal{A}} \max(0, v_j^R - v_i^R + \gamma),
\end{equation}
where $\gamma$ is a margin hyperparameter. As shown in~\cref{fig:sml_loss}, unlike BCE, the proposed objective enforces relative preference rather than absolute suppression, allowing related experts to retain meaningful activations while preserving specialization.

\subsection{Training Strategy}
\label{sec:3_4}
We train \mymethod\ in an end-to-end manner by integrating OME-LoRA into the backbone and optimize the model with a joint objective that combines contrastive learning and structural regularization.

\noindent \textbf{Architectural Integration.}
Instead of standard LoRA layers, we deploy OME-LoRA throughout the backbone, where each module consists of a shared adaptation path, modality-experts, and a learnable router. Such learnable parameters are maintained independently at each layer.

\noindent \textbf{Joint Optimization.}
The overall training objective $\mathcal{L}_{total}$ combines the primary contrastive objective with the proposed structural regularization losses:
\begin{equation}
\mathcal{L}_{total} = \mathcal{L}_{con} + \lambda_{ortho} \mathcal{L}_{ortho} + \lambda_{SML} \mathcal{L}_{SML},
\end{equation}
where $\lambda_{ortho}$ and $\lambda_{SML}$ control the strengths of the orthogonal penalty and Synergy Margin Loss, respectively. Crucially, both regularization terms $\mathcal{L}_{ortho}$ and $\mathcal{L}_{SML}$ are computed and averaged across all layers containing OME-LoRA modules. 

\begin{table*}[t]
\centering
\resizebox{.970\linewidth}{!}{%
    \begin{tabular}{lccccc}
    \toprule
      &   Image (36) &   Video (23)  &   Audio (12)  &   Audiovisual (10) &   All (81) \\
    \midrule
    \multicolumn{6}{c}{3B Models} \\
    \midrule
     Omni-Embed-Nemotron~\cite{xu2025omni} &  44.1 & 36.5 & 24.5 & 28.5 & 33.4 \\
     LCO-Emb~\cite{xiao2026scaling}   & 58.1 & \textbf{43.7} & 42.1 & 35.5 &  44.8 \\
     e5-omni~\cite{chen-etal-2026-e5}   & 64.8 & \underline{40.6} & 37.4 & 31.6 &  43.6 \\
     \cmidrule(lr){1-6}
     \baseline\ (Contrastive baseline) & \underline{66.8} & 40.4 & \underline{43.1} & \underline{40.4} &  \underline{47.7} \\
     \mymethod\ (Ours)  & \textbf{68.1} & 40.5 & \textbf{45.6} & \textbf{43.4} &  \textbf{49.4} \\
     \midrule
     \multicolumn{6}{c}{7B Models} \\
     \midrule
     OmniEmbed-v0.1~\cite{ma2025tevatron} & 44.2 & 35.0 & 31.6 & 29.0 &  35.0 \\
     OmniEmbed-Multivent~\cite{zhan2025magmar} &  51.8 &   40.2 &   39.5 & 34.4 & 41.5 \\
     LCO-Emb~\cite{xiao2026scaling}   & 61.6 & \textbf{45.5} & 45.2 & 35.5 &  46.9 \\
     e5-omni~\cite{chen-etal-2026-e5}   & \textbf{72.5} & \underline{44.3} & 44.3 & 42.3 &  \underline{50.9} \\
     WAVE~\cite{tang2026wave} & 42.7 & 39.8 & 33.4 & 38.1 & 38.5 \\
     \cmidrule(lr){1-6}
     \baseline\ (Contrastive baseline) & 71.3 & 40.2 & \underline{47.5} & \underline{43.9} & 50.7 \\ 
     \mymethod\ (Ours)  & \underline{71.7} & 41.4 &   \textbf{48.7} &     \textbf{44.6} &  \textbf{51.6} \\
    \bottomrule
    \end{tabular}
}
\caption{Comparison of omnimodal embedding models using Hit@1 (\%). The numbers in parentheses indicate the number of evaluation tasks for each modality and the total. Here, \baseline\ corresponds to a strong baseline trained with a standard LoRA setting. Across both 3B- and 7B-scale models, \mymethod\ achieves higher overall scores than previous methods, demonstrating well-balanced performance across diverse modalities, while also improving over our baseline. For each model scale, the best results are \textbf{bolded}, and the second-best are \underline{underlined}.}
\label{tab:method_comparison}
\end{table*}

\section{Experiment}
\label{sec:4_exp}
\noindent \textbf{Training Data.}
We curate an omnimodal data mixture of 1.12M samples spanning image, video, audio, and audiovisual modalities. The training mixture covers diverse tasks including retrieval, classification, and QA, with various modality compositions ranging from single-modality inputs to composed multimodal scenarios. Detailed dataset types and statistics are provided in~\cref{sec:train_data}.

\noindent \textbf{Model Variants.}
We build our \mymethod\ framework on top of pretrained Qwen-2.5-Omni~\cite{xu2025qwen25} and train both 3B and 7B scale variants. OME-LoRA modules are inserted into the LLM layers while keeping the remaining backbone parameters frozen. We additionally establish \baseline\ as a controlled baseline trained with the same data and optimization settings, using only a single shared LoRA and contrastive objective. 

\noindent \textbf{Implementation Details.}
Images and videos are processed using the native resolution scaling strategy of Qwen-2.5-VL~\cite{bai2025qwen2}, while videos are uniformly sampled into 8 frames and audio signals are resampled to 16~kHz. Models are trained for 1 epoch using the AdamW~\cite{loshchilov2017decoupled} optimizer with a cosine learning rate schedule, a peak learning rate of $10^{-4}$, a warmup ratio of 0.04, and a global batch size of 768. We employ GradCache~\cite{gao-etal-2021-scaling} to maintain a large global batch efficiently. For the contrastive loss, we use a temperature $\tau$ of 0.02. Consistent with prior methods~\cite{jiang2025vlmvec}, we pool on the EOS token position followed by L2 normalization to extract the final embeddings.

For OME-LoRA, we set the rank of the shared adaptation path to 8 and each modality-expert rank to 4. We use a synergy margin $\gamma$ of 0.2 and apply the proposed Progressive Synergy Routing (PSR) schedule with warmup and transition ratios of $t_{\text{warm}}=0.1$ and $t_{\text{blend}}=0.5$, respectively. Additional implementation details and hyperparameter configurations are provided in~\cref{sec:supple_implementation}.

\noindent \textbf{Evaluation Setup.}
We evaluate our framework on an extensive omnimodal benchmark consisting of 81 tasks spanning image, video, audio, and audiovisual modalities. Each modality group covers multiple embedding-oriented meta-tasks, such as retrieval, classification, and QA. For image and video evaluations, we adopt the MMEB-v1~\cite{jiang2025vlmvec} and MMEB-v2~\cite{meng2026vlmvecv} benchmarks, respectively, while further extending with additional audio and audiovisual benchmarks collected in this work. The complete evaluation dataset details are provided in~\cref{sec:eval_data}.

\noindent \textbf{Baselines.}
We compare \mymethod\ against recent Omni-MLLM-based embedding models, including OmniEmbed-v0.1~\cite{ma2025tevatron}, Omni-Embed-Nemotron~\cite{xu2025omni}, OmniEmbed-Multivent~\cite{zhan2025magmar}, WAVE~\cite{tang2026wave}, LCO-Emb~\cite{xiao2026scaling}, and e5-omni~\cite{chen-etal-2026-e5}. For a fair evaluation, we use officially released checkpoints and perform all evaluations under the unified protocol.

\noindent \textbf{Metrics.}
We use Hit@1 as the evaluation metric, measuring whether the ground-truth target is ranked at the top among candidates. We report both modality-specific average performance and the overall average score from a single run. The overall score is computed by averaging modality-specific scores, where each modality-specific score is obtained by averaging the corresponding meta-task scores. Here, each meta-task score is computed as the average of its associated benchmark scores. Detailed benchmark-level performance breakdowns are provided in~\cref{sec:per_dataset}.

\subsection{Main Results}
\noindent \textbf{Overall Performance.} 
In \cref{tab:method_comparison}, we compare our framework against recent Omni-MLLM embedding models. While some methods excel in specific modalities, such as LCO-Emb~\cite{xiao2026scaling} on video tasks and e5-omni~\cite{chen-etal-2026-e5} on image tasks, their strong performance is less consistent across modalities. In contrast, our contrastive baseline, \baseline, trained with a modality-shared LoRA, already achieves strong performance across all modalities at both the 3B and 7B scales. Building upon this baseline, \mymethod\ reaches the best overall scores of 49.4\% and 51.6\%, respectively. These improvements over \baseline\ are statistically significant across the 81 tasks in a paired t-test ($p=0.0044$ and $p=0.0189$).

\noindent
\textbf{Modality-wise Analysis.}
Breaking down the gains by modality, \mymethod\ improves most on audio and audiovisual tasks ($+2.5\%$ and $+3.0\%$ at the 3B scale), with more moderate margins on image and video. A similar pattern appears at the 7B scale, where e5-omni leads on image and video while \mymethod\ is stronger on audio and audiovisual tasks and attains a higher overall score. This yields more even coverage across all modalities.

\begin{table}[t]
\centering
\resizebox{.99\linewidth}{!}{%
\begin{tabular}{lcccc|ccccc}
\toprule
 & \multicolumn{2}{c}{OME-LoRA} & \multicolumn{2}{c|}{PSR} & & & & & \\
\cmidrule(lr){2-3} \cmidrule(lr){4-5}
id & S-Exp. & $\mathcal{L}_{ortho}$ & Soft & Prog. & I & V & A & AV & All \\
\midrule
(a) & - & - & - & - & 66.8 &   \underline{40.4} &   43.1 &  40.4 & 47.7 \\
\cmidrule(lr){1-10}
(b) & \checkmark & - & - & - & 67.6 &   39.6 &   44.9 &  \underline{42.6} & 48.7 \\
(c) & \checkmark & \checkmark & - & - & 67.4 &   39.9 &   \underline{45.2} &  41.7 & 48.6 \\
(d) & \checkmark & \checkmark & \checkmark & - & 67.3 &   39.9 &   45.0 &  42.1 & 48.6 \\
\cmidrule(lr){1-10}
(e) & \checkmark & - & \checkmark & \checkmark &  \textbf{68.2} &   40.1 &   \underline{45.2} &  \underline{42.6} & \underline{49.0} \\
(f) & \checkmark & \checkmark & \checkmark & \checkmark & \underline{68.1} &   \textbf{40.5} &   \textbf{45.6} &  \textbf{43.4} & \textbf{49.4} \\
\bottomrule
\end{tabular}
}
\caption{Component analysis of \mymethod\ framework. Starting from the \baseline\ baseline in (a), each row incorporates individual components of our design. The results show that both OME-LoRA and PSR contribute to improved and balanced performance over the baseline, with the full model achieving the best overall results.
}
\label{tab:component_analysis}
\end{table}

\subsection{Ablation Study}
We conduct a series of ablation experiments to analyze the individual contribution of each proposed component in our \mymethod\ framework. All results are reported using 3B-scale models, including both modality-specific and overall averaged scores.

\noindent \textbf{Component Analysis.}
As shown in~\cref{tab:component_analysis}, compared to \baseline\ baseline in (a), introducing a shared-expert structure (S-Exp.) with hard routing in (b) improves performance through structural decomposition. Since hard routing activates experts in isolation and leaves limited interaction for shared-expert decorrelation, the orthogonal penalty alone in (c) yields no additional gain over (b). Soft routing alone in (d) likewise maintains a comparable level, as it lets experts interact before acquiring modality priors, which blunts specialization. When applying Progressive Synergy Routing (PSR) with Synergy Margin Loss (SML) in (e), further improvements are obtained. Under this configuration, the orthogonal penalty provides a complementary gain in (f) (49.0 $\rightarrow$ 49.4), indicating that reducing redundancy becomes effective once multiple expert outputs are jointly integrated.

Beyond these overall numbers, the components introduced on top of the shared-expert structure also shape how experts behave: they yield clearer separation across pathways, progressive participation of absent-modality experts, and routing activations that avoid collapsing to hard assignments, which we further analyze in~\cref{sec:empirical_analysis}.

\noindent \textbf{Structural Efficiency.} 
We explore the impact of different architectural choices under matched parameter budgets. As shown in~\cref{tab:structure}, simply increasing the rank of a single shared path from 16 to 36 brings only marginal gains, with the rank 24 counterpart in (b) even leading to a slight drop. In contrast, our OME-LoRA decomposes the same total rank (24) into shared and expert paths, which brings improvements at an identical parameter count (44.9M) in (f). Applying PSR on top of this structure in (g) further boosts performance to 49.4$\%$. Notably, this surpasses the largest shared-only variant in (c) while using fewer trainable parameters, confirming that the gains stem from structured decomposition rather than increased capacity.

\begin{table}[t]
\centering
\resizebox{.99\linewidth}{!}{%
    \begin{tabular}{clccc|ccccc}
    \toprule
    id & Variant & Rank & Params. & PSR &  I & V & A & AV & All \\
    \midrule
    (a) & Shr & 16 & 29.9M & - &  66.8 & 40.4 & 43.1 &  40.4 & 47.7 \\
    (b) & Shr & 24 & 44.9M & - &  67.3 & 38.7 & 43.0 &  41.6 & 47.6 \\
    (c) & Shr & 36 & 59.9M & - &  67.3 & \textbf{40.7} & 43.8 &  41.2 & 48.2 \\
    \cmidrule(lr){1-10}
    (d) & Exp & 16 & 29.9M & - &  66.1 & 39.4 & 42.5 &  40.1 & 47.0 \\
    (e) & Exp & 16 & 33.3M & \checkmark & 65.3 & 39.5 & 42.1 &  40.5 & 46.9 \\
    \cmidrule(lr){1-10}
    (f) & Shr+Exp & 24 & 44.9M & - &  \underline{67.6} & 39.6 & \underline{44.9} &  \underline{42.6} & \underline{48.7} \\
    (g) & Shr+Exp & 24 & 48.3M & \checkmark & \textbf{68.1} & \underline{40.5} & \textbf{45.6} &  \textbf{43.4} & \textbf{49.4} \\
    \bottomrule
    \end{tabular}
}
\caption{Analysis of the adaptation structure. `Shr' denotes a single shared LoRA path (\baseline) with increasing rank. `Exp' denotes modality-expert paths without a shared LoRA path, and `Shr+Exp' denotes our OME-LoRA architecture. Neither path alone is sufficient: scaling the shared path yields limited gains, and removing it degrades performance further.
\vspace{2mm}
}
\label{tab:structure}
\end{table}

\begin{table}[t]
\centering
\resizebox{.99\linewidth}{!}{%
    \begin{tabular}{lccc|ccccc}
    \toprule
     & \multicolumn{3}{c|}{Routing Strategy (PSR)} & & & & & \\
    \cmidrule(lr){2-4}
    id & Router & Progress & loss & I & V & A & AV & All \\
    \midrule
    (a) & Hard & - & - & 67.4 & 39.9 & 45.2 & 41.7 & 48.6 \\
    \cmidrule(lr){1-9}
    (b) & Soft & - & BCE & 67.3 & 39.9 & 45.0 & \underline{42.1} & 48.6 \\
    (c) & Soft  & - & SML & \underline{67.6} & \underline{40.4} & 45.1 & 41.9 & \underline{48.8} \\
    \cmidrule(lr){1-9}
    (d) & Soft  & \checkmark & BCE & \underline{67.6} & 39.6 & \textbf{46.0} & 41.9 & \underline{48.8} \\
    (e) & Soft  & \checkmark & SML & \textbf{68.1} & \textbf{40.5} & \underline{45.6} & \textbf{43.4} & \textbf{49.4} \\
    \bottomrule
    \end{tabular}

}
\caption{Analysis on Progressive Synergy Routing (PSR) strategy. We evaluate the impact of routing types, progressive blending, and router loss functions in the OME-LoRA layers. We highlight that combining a progressive blending of routing weights with the Synergy Margin Loss (SML) is crucial for both expert specialization and cross-modal collaboration among experts.
}
\label{tab:psr_ablation}
\end{table}

\noindent \textbf{Role of Shared LoRA.} 
We next evaluate the necessity of the shared LoRA path within the OME-LoRA architecture. As shown in~\cref{tab:structure}, omitting the shared path leads to a substantial performance drop under both hard routing (f vs. d) and with PSR integrated (g vs. e) cases. More importantly, expert-only adaptation in (d) falls even below the shared-only baseline in (a) at the same parameter count (47.0 vs. 47.7 at 29.9M). This suggests that the shared path provides a unified cross-modal space on which experts build their specialization, and that establishing it matters more than allocating the same capacity to experts alone.

\noindent \textbf{Role of Progressive Synergy Routing.}
We further investigate the impact of our routing strategies, as summarized in~\cref{tab:psr_ablation}. Compared to the hard routing baseline in (a), switching to soft routing with BCE loss in (b) provides limited to no improvement. However, across both standard soft routing (b vs. c) and progressive blending configurations (d vs. e), replacing BCE with the Synergy Margin Loss (SML) consistently improves performance. Our final framework in (e), which integrates both progressive blending and SML, achieves the best overall performance of 49.4\%. This demonstrates that the two components, progressive blending with synergy margin loss in PSR, complement each other to deliver the superior results.

\begin{figure}[t]
  \centering
   \includegraphics[width=0.99\columnwidth]{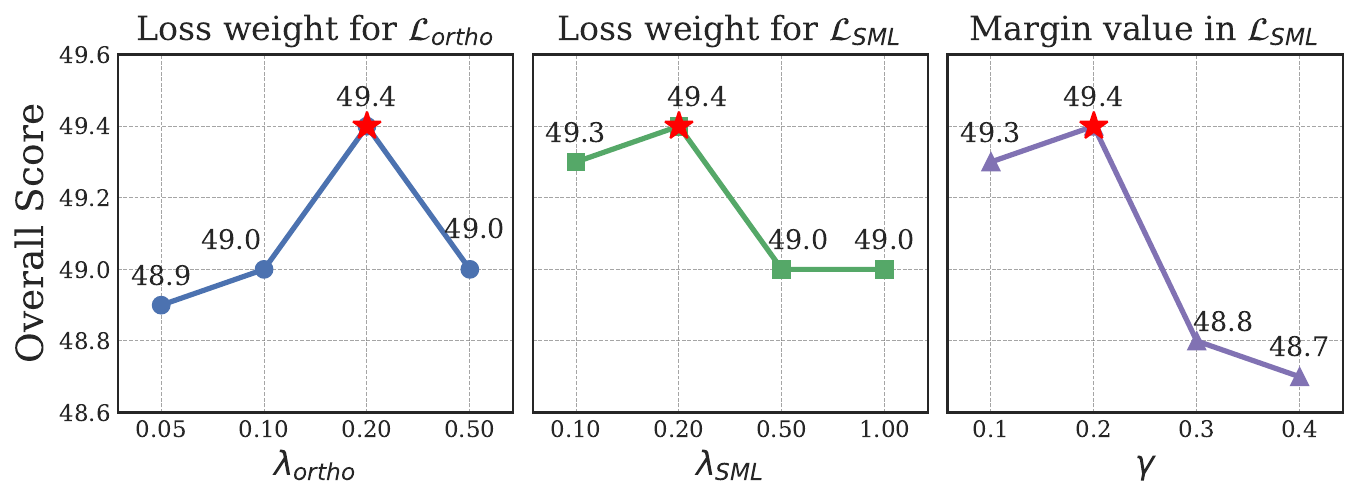}
    \caption{Sensitivity analysis of hyperparameters on the orthogonal penalty weight $\lambda_{ortho}$, the synergy margin loss weight $\lambda_{SML}$, and the margin parameter $\gamma$. Red stars indicate the optimal choice in our final model.}
   \label{fig:sensitivity}
\end{figure}

\begin{figure}[t]
  \centering
   \includegraphics[width=0.99\columnwidth]{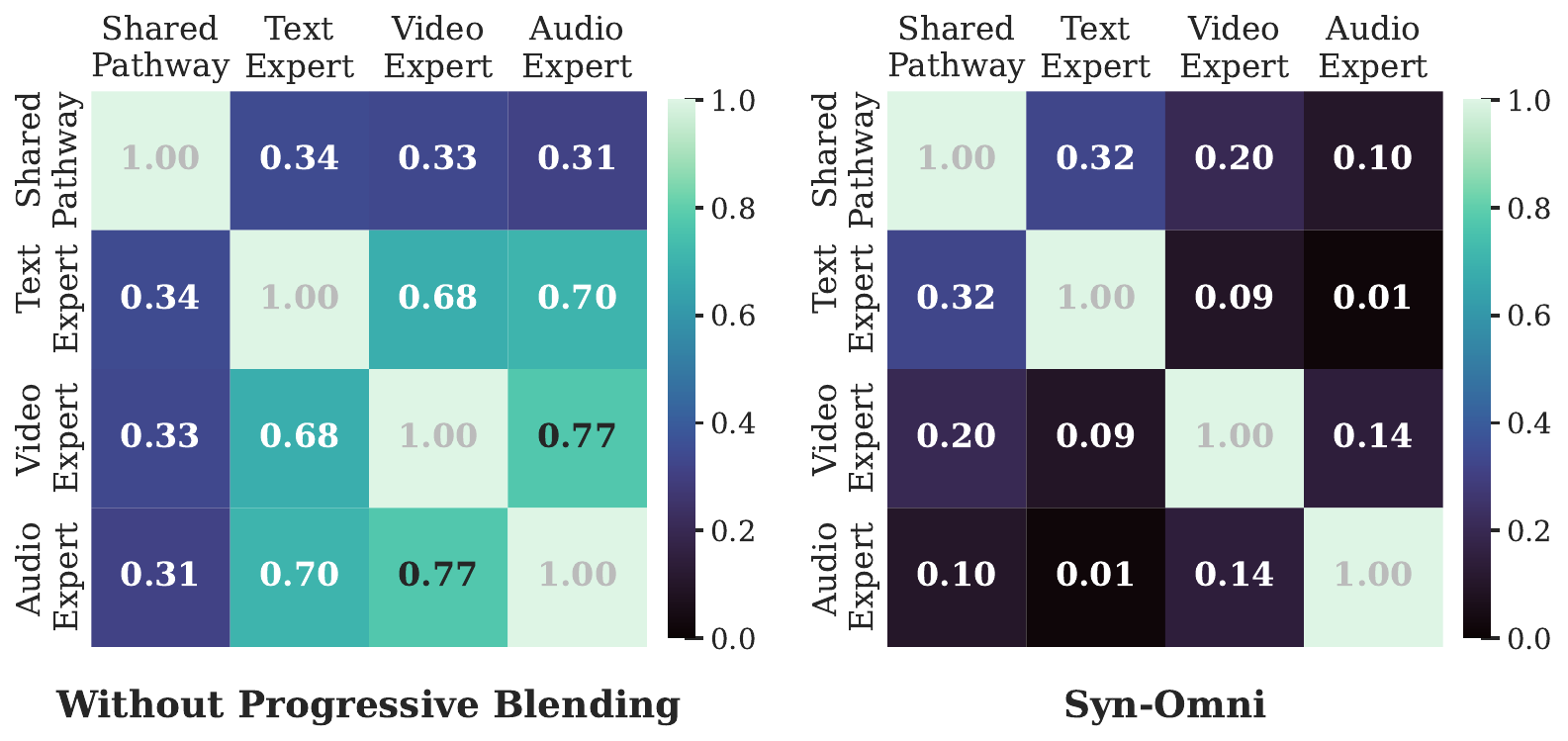}
    \caption{Pairwise cosine similarity between OME-LoRA pathways. Without progressive blending (left), high cross-path similarities reveal redundancy across different pathways. In contrast, \mymethod\ (right), which differs only in applying progressive blending, produces significantly lower similarities. This indicates effective decoupling and clear specialization across pathways.
    }
   \label{fig:similarity}
\end{figure}

\noindent\textbf{Sensitivity Analysis.} 
We investigate the sensitivity of three key hyperparameters in \mymethod: the orthogonal penalty weight ($\lambda_{ortho}$), the synergy margin loss weight ($\lambda_{SML}$), and the margin parameter ($\gamma$). As shown in~\cref{fig:sensitivity}, the best overall score is achieved with $\lambda_{ortho}=0.20$, $\lambda_{SML}=0.20$, and $\gamma=0.2$. The performance remains relatively stable around these values, while larger margin values lead to a noticeable performance drop.

\begin{figure}[t]
  \centering
   \includegraphics[width=0.99\columnwidth]{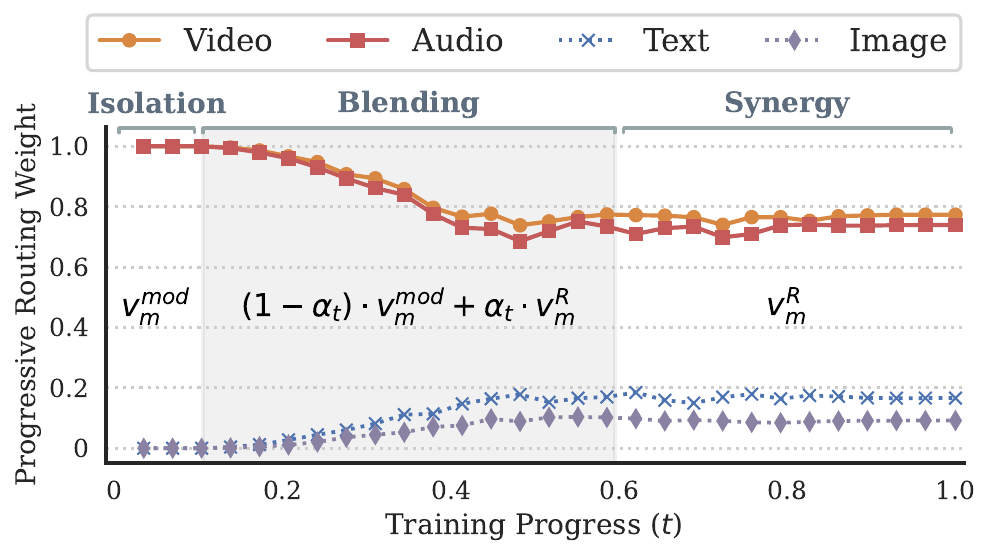}
    \caption{Routing dynamics of PSR under video-audio inputs. Present-modality experts retain dominant routing weights, while absent-modality experts gradually acquire non-zero activations through progressive blending. This indicates the emergence of synergistic cross-modal interactions along with modality specialization.}
   \label{fig:progress_routing}
\end{figure}

\subsection{Empirical Analysis}
\label{sec:empirical_analysis}
To better understand how \mymethod\ structures omnimodal representations, we analyze the behaviors of its two key components, OME-LoRA and PSR, on VALOR-32k~\cite{10721284} evaluation samples with text, video, and audio inputs. Results are computed from final-layer signals averaged across samples and submodules. 

\noindent \textbf{Decoupling in OME-LoRA.}
As shown in~\cref{fig:similarity}, we measure pairwise cosine similarity between the flattened hidden states of the shared pathway and modality-experts using text-video-audio inputs. Without applying progressive blending (left), relying solely on soft routing fails to induce clear separation, evidenced by high cross-path similarities both between the shared pathway and experts, and across experts. In contrast, \mymethod\ (right) produces substantially lower cross-path similarities, indicating clear modality specialization across paths in the OME-LoRA architecture. 

\noindent \textbf{Routing Dynamics in PSR.}
We analyze the evolution of PSR routing weights under video-audio inputs to demonstrate the emergence of cross-modal interactions during training in~\cref{fig:progress_routing}. Under progressive blending, present-modality experts (video and audio) maintain dominant routing weights. In contrast, absent-modality experts (text and image), initially suppressed to zero under hard routing scheme, gradually acquire non-zero weights by introducing soft routing scheme. This behavior demonstrates that PSR enables cross-modal collaboration while preserving modality specialization. We further observe that SML drives this synergistic behavior, whereas BCE loss completely suppresses absent-modality experts, as shown in~\cref{fig:supple_heatmap}.

\section{Conclusion}
We propose \mymethod, a structured omnimodal adaptation framework that effectively learns a unified representation space. It leverages OME-LoRA to decouple modality-shared and specialized parameters. In addition, Progressive Synergy Routing (PSR) shifts the model from isolated specialization toward dynamic cross-modal collaboration. Across 81 diverse tasks, \mymethod\ outperforms existing baselines, achieving robust and well-balanced performance across image, video, audio, and audiovisual modalities.

\section*{Limitations}
Our framework introduces additional inference overhead through dynamic routing in the OME-LoRA layer. While standard LoRA weights can be merged into the frozen backbone after training, our input-dependent routing prevents this for the modality-experts, although the shared pathway remains mergeable. Specifically, the increased overhead due to the expert pathways active at inference amounts to 17--18$\%$ higher latency with less than 1$\%$ additional GPU memory, as reported in~\cref{tab:runtime_analysis}. This cost is most visible in latency-sensitive settings such as real-time query encoding.

\section*{Ethical Considerations}
Our framework may inherit systemic biases from large-scale training data, leading to skewed cross-modal retrieval behaviors. Additionally, such unified omnimodal embeddings could be misused for unauthorized surveillance or identity linkage, necessitating strict deployment oversight.

\bibliography{custom}

@inproceedings{
zhang2023adaptive,
title={Adaptive Budget Allocation for Parameter-Efficient Fine-Tuning},
author={Qingru Zhang and Minshuo Chen and Alexander Bukharin and Pengcheng He and Yu Cheng and Weizhu Chen and Tuo Zhao},
booktitle={The Eleventh International Conference on Learning Representations},
year={2023}
}

@inproceedings{bengio2009curriculum,
  title={Curriculum learning},
  author={Bengio, Yoshua and Louradour, J{\'e}r{\^o}me and Collobert, Ronan and Weston, Jason},
  booktitle={Proceedings of the 26th annual international conference on machine learning},
  pages={41--48},
  year={2009}
}

@inproceedings{
li2025stella,
title={Stel{LA}: Subspace Learning in Low-rank Adaptation using Stiefel Manifold},
author={Zhizhong Li and Sina Sajadmanesh and Jingtao Li and Lingjuan Lyu},
booktitle={The Thirty-ninth Annual Conference on Neural Information Processing Systems},
year={2025}
}

@article{luo2024moelora,
  title={Moelora: Contrastive learning guided mixture of experts on parameter-efficient fine-tuning for large language models},
  author={Luo, Tongxu and Lei, Jiahe and Lei, Fangyu and Liu, Weihao and He, Shizhu and Zhao, Jun and Liu, Kang},
  journal={arXiv preprint arXiv:2402.12851},
  year={2024}
}

@inproceedings{
jiang2025vlmvec,
title={{VLM}2Vec: Training Vision-Language Models for Massive Multimodal Embedding Tasks},
author={Ziyan Jiang and Rui Meng and Xinyi Yang and Semih Yavuz and Yingbo Zhou and Wenhu Chen},
booktitle={The Thirteenth International Conference on Learning Representations},
year={2025},
url={https://openreview.net/forum?id=TE0KOzWYAF}
}

@inproceedings{das2017visual,
  title={Visual dialog},
  author={Das, Abhishek and Kottur, Satwik and Gupta, Khushi and Singh, Avi and Yadav, Deshraj and Moura, Jos{\'e} MF and Parikh, Devi and Batra, Dhruv},
  booktitle={Proceedings of the IEEE conference on computer vision and pattern recognition},
  pages={326--335},
  year={2017}
}

@inproceedings{liu2021image,
  title={Image retrieval on real-life images with pre-trained vision-and-language models},
  author={Liu, Zheyuan and Rodriguez-Opazo, Cristian and Teney, Damien and Gould, Stephen},
  booktitle={Proceedings of the IEEE/CVF international conference on computer vision},
  pages={2125--2134},
  year={2021}
}

@inproceedings{
fu2023dreamsim,
title={DreamSim: Learning New Dimensions of Human Visual Similarity using Synthetic Data},
author={Stephanie Fu and Netanel Yakir Tamir and Shobhita Sundaram and Lucy Chai and Richard Zhang and Tali Dekel and Phillip Isola},
booktitle={Thirty-seventh Conference on Neural Information Processing Systems},
year={2023},
url={https://openreview.net/forum?id=DEiNSfh1k7}
}

@inproceedings{hendrycks2021many,
  title={The many faces of robustness: A critical analysis of out-of-distribution generalization},
  author={Hendrycks, Dan and Basart, Steven and Mu, Norman and Kadavath, Saurav and Wang, Frank and Dorundo, Evan and Desai, Rahul and Zhu, Tyler and Parajuli, Samyak and Guo, Mike and Song, Dawn and Steinhardt, Jacob and Gilmer, Justin},
  booktitle={Proceedings of the IEEE/CVF international conference on computer vision},
  pages={8340--8349},
  year={2021}
}

@inproceedings{barbu2019objectnet,
 author = {Barbu, Andrei and Mayo, David and Alverio, Julian and Luo, William and Wang, Christopher and Gutfreund, Dan and Tenenbaum, Josh and Katz, Boris},
 booktitle = {Advances in Neural Information Processing Systems},
 title = {ObjectNet: A large-scale bias-controlled dataset for pushing the limits of object recognition models},
 volume = {32},
 pages={9453--9463},
 year = {2019}
}

@InProceedings{radford2021learning,
  title = 	 {Learning Transferable Visual Models From Natural Language Supervision},
  author =       {Radford, Alec and Kim, Jong Wook and Hallacy, Chris and Ramesh, Aditya and Goh, Gabriel and Agarwal, Sandhini and Sastry, Girish and Askell, Amanda and Mishkin, Pamela and Clark, Jack and Krueger, Gretchen and Sutskever, Ilya},
  booktitle = 	 {Proceedings of the 38th International Conference on Machine Learning},
  pages = 	 {8748--8763},
  year = 	 {2021},
  volume = 	 {139}
}

@inproceedings{hendrycks2021natural,
  title={Natural adversarial examples},
  author={Hendrycks, Dan and Zhao, Kevin and Basart, Steven and Steinhardt, Jacob and Song, Dawn},
  booktitle={Proceedings of the IEEE/CVF conference on computer vision and pattern recognition},
  pages={15262--15271},
  year={2021}
}

@ARTICLE{zhou2017places,
  author={Zhou, Bolei and Lapedriza, Agata and Khosla, Aditya and Oliva, Aude and Torralba, Antonio},
  journal={IEEE Transactions on Pattern Analysis and Machine Intelligence}, 
  title={Places: A 10 Million Image Database for Scene Recognition}, 
  year={2018},
  volume={40},
  number={6},
  pages={1452--1464},
  doi={10.1109/TPAMI.2017.2723009}
}

@inproceedings{ye2024cat,
  title={Cat: Enhancing multimodal large language model to answer questions in dynamic audio-visual scenarios},
  author={Ye, Qilang and Yu, Zitong and Shao, Rui and Xie, Xinyu and Torr, Philip and Cao, Xiaochun},
  booktitle={European Conference on Computer Vision},
  pages={146--164},
  year={2024},
  organization={Springer}
}

@inproceedings{wang2024internvideo2,
author = {Wang, Yi and Li, Kunchang and Li, Xinhao and Yu, Jiashuo and He, Yinan and Chen, Guo and Pei, Baoqi and Zheng, Rongkun and Wang, Zun and Shi, Yansong and Jiang, Tianxiang and Li, Songze and Xu, Jilan and Zhang, Hongjie and Huang, Yifei and Qiao, Yu and Wang, Yali and Wang, Limin},
title = {InternVideo2: Scaling Foundation Models for Multimodal Video Understanding},
year = {2024},
booktitle={European conference on computer vision},
pages = {396--416},
}

@inproceedings{han2025shot2story,
  title={Shot2story: A new benchmark for comprehensive understanding of multi-shot videos},
  author={Han, Mingfei and Yang, Linjie and Chang, Xiaojun and Yao, Lina and Wang, Heng},
  booktitle={International Conference on Learning Representations},
  year={2025}
}

@inproceedings{
chen2023vast,
title={{VAST}: A Vision-Audio-Subtitle-Text Omni-Modality Foundation Model and Dataset},
author={Sihan Chen and Handong Li and Qunbo Wang and Zijia Zhao and Mingzhen Sun and Xinxin Zhu and Jing Liu},
booktitle={Thirty-seventh Conference on Neural Information Processing Systems},
year={2023},
url={https://openreview.net/forum?id=scYa9DYUAy}
}

@article{agostinelli2023musiclm,
  title={Musiclm: Generating music from text},
  author={Agostinelli, Andrea and Denk, Timo I and Borsos, Zal{\'a}n and Engel, Jesse and Verzetti, Mauro and Caillon, Antoine and Huang, Qingqing and Jansen, Aren and Roberts, Adam and Tagliasacchi, Marco and Sharifi, Matt and Zeghidour, Neil and Frank, Christian},
  journal={arXiv preprint arXiv:2301.11325},
  year={2023}
}

@inproceedings{gemmeke2017audio,
  title={Audio set: An ontology and human-labeled dataset for audio events},
  author={Gemmeke, Jort F and Ellis, Daniel PW and Freedman, Dylan and Jansen, Aren and Lawrence, Wade and Moore, R Channing and Plakal, Manoj and Ritter, Marvin},
  booktitle={2017 IEEE international conference on acoustics, speech and signal processing (ICASSP)},
  pages={776--780},
  year={2017},
  organization={IEEE}
}

@article{mei2024wavcaps,
  title={Wavcaps: A chatgpt-assisted weakly-labelled audio captioning dataset for audio-language multimodal research},
  author={Mei, Xinhao and Meng, Chutong and Liu, Haohe and Kong, Qiuqiang and Ko, Tom and Zhao, Chengqi and Plumbley, Mark D and Zou, Yuexian and Wang, Wenwu},
  journal={IEEE/ACM Transactions on Audio, Speech, and Language Processing},
  volume={32},
  pages={3339--3354},
  year={2024},
  publisher={IEEE}
}

@inproceedings{
loshchilov2017decoupled,
title={Decoupled Weight Decay Regularization},
author={Ilya Loshchilov and Frank Hutter},
booktitle={International Conference on Learning Representations},
year={2019},
url={https://openreview.net/forum?id=Bkg6RiCqY7},
}

@article{zhan2025magmar,
  title={MAGMaR Shared Task System Description: Video Retrieval with OmniEmbed},
  author={Zhan, Jiaqi Samantha and Zhang, Crystina and Zhuang, Shengyao and Ma, Xueguang and Lin, Jimmy},
  journal={arXiv preprint arXiv:2506.09409},
  year={2025}
}

@inproceedings{
ma2026mmar,
title={{MMAR}: A Challenging Benchmark for Deep Reasoning in Speech, Audio, Music, and Their Mix},
author={Ziyang Ma and Yinghao Ma and Yanqiao Zhu and Chen Yang and Yi-Wen Chao and Ruiyang Xu and Wenxi Chen and Yuanzhe Chen and Zhuo Chen and Jian Cong and Kai Li and Keliang Li and Siyou Li and Xinfeng Li and Xiquan Li and Zheng Lian and Yuzhe Liang and Minghao Liu and Zhikang Niu and Tianrui Wang and Yuping Wang and Yuxuan Wang and Yihao Wu and Guanrou Yang and Jianwei Yu and Ruibin Yuan and Zhisheng Zheng and Ziya Zhou and Haina Zhu and Wei Xue and Emmanouil Benetos and Kai Yu and EngSiong Chng and Xie Chen},
booktitle={The Thirty-ninth Annual Conference on Neural Information Processing Systems Datasets and Benchmarks Track},
year={2025},
url={https://openreview.net/forum?id=fgmrBJemlQ}
}

@article{zhou2025daily,
  title={Daily-omni: Towards audio-visual reasoning with temporal alignment across modalities},
  author={Zhou, Ziwei and Wang, Rui and Wu, Zuxuan and Jiang, Yu-Gang},
  journal={arXiv preprint arXiv:2505.17862},
  year={2025}
}

@inproceedings{sung-bin2025avhbench,
  title={Avhbench: A cross-modal hallucination benchmark for audio-visual large language models},
  author={Kim, Sung-Bin and Oh, Hyun-Bin and Lee, Jung-Mok and Senocak, Arda and Chung, Joon Son and Oh, Tae-Hyun},
  booktitle={International Conference on Learning Representations},
  year={2025},
  url={https://openreview.net/forum?id=jTEKTdI3K9}
}

@inproceedings{tian2018audio,
  title={Audio-visual event localization in unconstrained videos},
  author={Tian, Yapeng and Shi, Jing and Li, Bochen and Duan, Zhiyao and Xu, Chenliang},
  booktitle={Proceedings of the European conference on computer vision (ECCV)},
  pages={247--263},
  year={2018}
}

@inproceedings{
sakshi2025mmau,
title={{MMAU}: A Massive Multi-Task Audio Understanding and Reasoning Benchmark},
author={S Sakshi and Utkarsh Tyagi and Sonal Kumar and Ashish Seth and Ramaneswaran Selvakumar and Oriol Nieto and Ramani Duraiswami and Sreyan Ghosh and Dinesh Manocha},
booktitle={The Thirteenth International Conference on Learning Representations},
year={2025},
url={https://openreview.net/forum?id=TeVAZXr3yv}
}

@inproceedings{yang2022avqa,
  title={Avqa: A dataset for audio-visual question answering on videos},
  author={Yang, Pinci and Wang, Xin and Duan, Xuguang and Chen, Hong and Hou, Runze and Jin, Cong and Zhu, Wenwu},
  booktitle={Proceedings of the 30th ACM international conference on multimedia},
  pages={3480--3491},
  year={2022}
}

@inproceedings{engel2017neural,
  title={Neural audio synthesis of musical notes with wavenet autoencoders},
  author={Engel, Jesse and Resnick, Cinjon and Roberts, Adam and Dieleman, Sander and Norouzi, Mohammad and Eck, Douglas and Simonyan, Karen},
  booktitle={International conference on machine learning},
  pages={1068--1077},
  year={2017},
  organization={PMLR}
}

@article{tzanetakis2002musical,
  title={Musical genre classification of audio signals},
  author={Tzanetakis, George and Cook, Perry},
  journal={IEEE Transactions on speech and audio processing},
  volume={10},
  number={5},
  pages={293--302},
  year={2002},
  publisher={IEEE}
}

@inproceedings{piczak2015esc,
  title={ESC: Dataset for environmental sound classification},
  author={Piczak, Karol J},
  booktitle={Proceedings of the 23rd ACM international conference on Multimedia},
  pages={1015--1018},
  year={2015}
}

@article{koepke2022audio,
author = {Koepke, A. Sophia and Oncescu, Andreea-Maria and Henriques, Jo{\~a}o F. and Akata, Zeynep and Albanie, Samuel},
title = {Audio Retrieval With Natural Language Queries: A Benchmark Study},
year = {2023},
publisher = {IEEE Press},
volume = {25},
doi = {10.1109/TMM.2022.3149712},
journal={IEEE Transactions on Multimedia},
pages = {2675--2685}
}

@ARTICLE{10721284,
  author={Liu, Jing and Chen, Sihan and He, Xingjian and Guo, Longteng and Zhu, Xinxin and Wang, Weining and Tang, Jinhui},
  journal={IEEE Transactions on Pattern Analysis and Machine Intelligence}, 
  title={VALOR: Vision-Audio-Language Omni-Perception Pretraining Model and Dataset}, 
  year={2025},
  volume={47},
  number={2},
  pages={708--724},
  doi={10.1109/TPAMI.2024.3479776}}

@inproceedings{bac89bfb28074b369efe4238b0048d5e,
  title={Clotho: An audio captioning dataset},
  author={Drossos, Konstantinos and Lipping, Samuel and Virtanen, Tuomas},
  booktitle={{IEEE} International Conference on Acoustics, Speech and Signal Processing (ICASSP)},
  pages={736--740},
  year={2020}
}

@inproceedings{
wu2025learning,
title={Learning Fine-Grained Representations through Textual Token Disentanglement in Composed Video Retrieval},
author={Yue Wu and Zhaobo Qi and Yiling Wu and Junshu Sun and Yaowei Wang and Shuhui Wang},
booktitle={The Thirteenth International Conference on Learning Representations},
year={2025},
url={https://openreview.net/forum?id=wGa2plE8ka}
}

@inproceedings{chang2022webqa,
  title={Webqa: Multihop and multimodal qa},
  author={Chang, Yingshan and Narang, Mridu and Suzuki, Hisami and Cao, Guihong and Gao, Jianfeng and Bisk, Yonatan},
  booktitle={Proceedings of the IEEE/CVF conference on computer vision and pattern recognition},
  pages={16495--16504},
  year={2022}
}

@inproceedings{krishna2017dense,
  title={Dense-captioning events in videos},
  author={Krishna, Ranjay and Hata, Kenji and Ren, Frederic and Fei-Fei, Li and Carlos Niebles, Juan},
  booktitle={Proceedings of the IEEE international conference on computer vision},
  pages={706--715},
  year={2017}
}

@inproceedings{anne2017localizing,
  title={Localizing moments in video with natural language},
  author={Hendricks, Lisa Anne and Wang, Oliver and Shechtman, Eli and Sivic, Josef and Darrell, Trevor and Russell, Bryan},
  booktitle={Proceedings of the IEEE international conference on computer vision},
  pages={5803--5812},
  year={2017}
}

@inproceedings{xu2016msr,
  title={Msr-vtt: A large video description dataset for bridging video and language},
  author={Xu, Jun and Mei, Tao and Yao, Ting and Rui, Yong},
  booktitle={Proceedings of the IEEE conference on computer vision and pattern recognition},
  pages={5288--5296},
  year={2016}
}

@inproceedings{caba2015activitynet,
  title={Activitynet: A large-scale video benchmark for human activity understanding},
  author={Caba Heilbron, Fabian and Escorcia, Victor and Ghanem, Bernard and Carlos Niebles, Juan},
  booktitle={Proceedings of the ieee conference on computer vision and pattern recognition},
  pages={961--970},
  year={2015}
}

@inproceedings{bolya2026perception,
 author = {Bolya, Daniel and Huang, Po-Yao and Sun, Peize and Cho, Jang Hyun and Madotto, Andrea and Wei, Chen and Ma, Tengyu and Zhi, Jiale and Rajasegaran, Jathushan and Bangalath, Hanoona and Wang, Junke and Monteiro, Marco and Xu, Hu and Dong, Shiyu and Ravi, Nikhila and Li, Shang-Wen and Dollar, Piotr and Feichtenhofer, Christoph},
 booktitle = {Advances in Neural Information Processing Systems},
 pages = {60884--60937},
 title = {Perception Encoder: The best visual embeddings are not at the output of the network},
 volume = {38},
 year = {2025}
}

@inproceedings{lin2014microsoft,
  title={Microsoft coco: Common objects in context},
  author={Lin, Tsung-Yi and Maire, Michael and Belongie, Serge and Hays, James and Perona, Pietro and Ramanan, Deva and Doll{\'a}r, Piotr and Zitnick, C Lawrence},
  booktitle={European conference on computer vision},
  pages={740--755},
  year={2014},
  organization={Springer}
}

@inproceedings{hudson2019gqa,
  title={Gqa: A new dataset for real-world visual reasoning and compositional question answering},
  author={Hudson, Drew A and Manning, Christopher D},
  booktitle={Proceedings of the IEEE/CVF conference on computer vision and pattern recognition},
  pages={6700--6709},
  year={2019}
}

@inproceedings{singh2019towards,
  title={Towards vqa models that can read},
  author={Singh, Amanpreet and Natarajan, Vivek and Shah, Meet and Jiang, Yu and Chen, Xinlei and Batra, Dhruv and Parikh, Devi and Rohrbach, Marcus},
  booktitle={Proceedings of the IEEE/CVF conference on computer vision and pattern recognition},
  pages={8317--8326},
  year={2019}
}

@inproceedings{kuehne2014language,
  title={The language of actions: Recovering the syntax and semantics of goal-directed human activities},
  author={Kuehne, Hilde and Arslan, Ali and Serre, Thomas},
  booktitle={Proceedings of the IEEE conference on computer vision and pattern recognition},
  pages={780--787},
  year={2014}
}

@article{soomro2012ucf101,
  title={Ucf101: A dataset of 101 human actions classes from videos in the wild},
  author={Soomro, Khurram and Zamir, Amir Roshan and Shah, Mubarak},
  journal={arXiv preprint arXiv:1212.0402},
  year={2012}
}

@InProceedings{goyal2017something,
author = {Goyal, Raghav and Ebrahimi Kahou, Samira and Michalski, Vincent and Materzynska, Joanna and Westphal, Susanne and Kim, Heuna and Haenel, Valentin and Fruend, Ingo and Yianilos, Peter and Mueller-Freitag, Moritz and Hoppe, Florian and Thurau, Christian and Bax, Ingo and Memisevic, Roland},
title = {The "Something Something" Video Database for Learning and Evaluating Visual Common Sense},
booktitle = {Proceedings of the IEEE International Conference on Computer Vision (ICCV)},
year = {2017}
}

@inproceedings{kuehne2011hmdb,
  title={HMDB: a large video database for human motion recognition},
  author={Kuehne, Hildegard and Jhuang, Hueihan and Garrote, Est{\'\i}baliz and Poggio, Tomaso and Serre, Thomas},
  booktitle={2011 International conference on computer vision},
  pages={2556--2563},
  year={2011},
  organization={IEEE}
}

@article{carreira2019short,
  title={A short note on the kinetics-700 human action dataset},
  author={Carreira, Joao and Noland, Eric and Hillier, Chloe and Zisserman, Andrew},
  journal={arXiv preprint arXiv:1907.06987},
  year={2019}
}

@inproceedings{hu2023open,
  title={Open-domain visual entity recognition: Towards recognizing millions of wikipedia entities},
  author={Hu, Hexiang and Luan, Yi and Chen, Yang and Khandelwal, Urvashi and Joshi, Mandar and Lee, Kenton and Toutanova, Kristina and Chang, Ming-Wei},
  booktitle={Proceedings of the IEEE/CVF International Conference on Computer Vision},
  pages={12065--12075},
  year={2023}
}

@inproceedings{wu2021fashion,
  title={Fashion iq: A new dataset towards retrieving images by natural language feedback},
  author={Wu, Hui and Gao, Yupeng and Guo, Xiaoxiao and Al-Halah, Ziad and Rennie, Steven and Grauman, Kristen and Feris, Rogerio},
  booktitle={Proceedings of the IEEE/CVF Conference on computer vision and pattern recognition},
  pages={11307--11317},
  year={2021}
}

@inproceedings{gurari2018vizwiz,
  title={Vizwiz grand challenge: Answering visual questions from blind people},
  author={Gurari, Danna and Li, Qing and Stangl, Abigale J and Guo, Anhong and Lin, Chi and Grauman, Kristen and Luo, Jiebo and Bigham, Jeffrey P},
  booktitle={Proceedings of the IEEE conference on computer vision and pattern recognition},
  pages={3608--3617},
  year={2018}
}

@inproceedings{lu2022learn,
 author = {Lu, Pan and Mishra, Swaroop and Xia, Tanglin and Qiu, Liang and Chang, Kai-Wei and Zhu, Song-Chun and Tafjord, Oyvind and Clark, Peter and Kalyan, Ashwin},
 booktitle = {Advances in Neural Information Processing Systems},
 pages = {2507--2521},
 title = {Learn to Explain: Multimodal Reasoning via Thought Chains for Science Question Answering},
 volume = {35},
 year = {2022}
}

@inproceedings{zhu2016visual7w,
  title={Visual7w: Grounded question answering in images},
  author={Zhu, Yuke and Groth, Oliver and Bernstein, Michael and Fei-Fei, Li},
  booktitle={Proceedings of the IEEE conference on computer vision and pattern recognition},
  pages={4995--5004},
  year={2016}
}

@inproceedings{mathew2022infographicvqa,
  title={Infographicvqa},
  author={Mathew, Minesh and Bagal, Viraj and Tito, Rub{\`e}n and Karatzas, Dimosthenis and Valveny, Ernest and Jawahar, CV},
  booktitle={Proceedings of the IEEE/CVF Winter Conference on Applications of Computer Vision},
  pages={1697--1706},
  year={2022}
}

@inproceedings{mathew2021docvqa,
  title={Docvqa: A dataset for vqa on document images},
  author={Mathew, Minesh and Karatzas, Dimosthenis and Jawahar, CV},
  booktitle={Proceedings of the IEEE/CVF winter conference on applications of computer vision},
  pages={2200--2209},
  year={2021}
}

@inproceedings{schwenk2022okvqa,
  title={A-okvqa: A benchmark for visual question answering using world knowledge},
  author={Schwenk, Dustin and Khandelwal, Apoorv and Clark, Christopher and Marino, Kenneth and Mottaghi, Roozbeh},
  booktitle={European conference on computer vision},
  pages={146--162},
  year={2022},
  organization={Springer}
}

@inproceedings{xiao2010sun,
  title={Sun database: Large-scale scene recognition from abbey to zoo},
  author={Xiao, Jianxiong and Hays, James and Ehinger, Krista A and Oliva, Aude and Torralba, Antonio},
  booktitle={2010 IEEE computer society conference on computer vision and pattern recognition},
  pages={3485--3492},
  year={2010},
  organization={IEEE}
}

@inproceedings{marino2019ok,
  title={Ok-vqa: A visual question answering benchmark requiring external knowledge},
  author={Marino, Kenneth and Rastegari, Mohammad and Farhadi, Ali and Mottaghi, Roozbeh},
  booktitle={Proceedings of the IEEE/cvf conference on computer vision and pattern recognition},
  pages={3195--3204},
  year={2019}
}

@article{everingham2015pascal,
  title={The pascal visual object classes challenge: A retrospective},
  author={Everingham, Mark and Eslami, S. M. Ali and Van Gool, Luc and Williams, Christopher KI and Winn, John and Zisserman, Andrew},
  journal={International journal of computer vision},
  volume={111},
  number={1},
  pages={98--136},
  year={2015},
  publisher={Springer}
}

@inproceedings{kiela2020hateful,
 author = {Kiela, Douwe and Firooz, Hamed and Mohan, Aravind and Goswami, Vedanuj and Singh, Amanpreet and Ringshia, Pratik and Testuggine, Davide},
 booktitle = {Advances in Neural Information Processing Systems},
 pages = {2611--2624},
 title = {The Hateful Memes Challenge: Detecting Hate Speech in Multimodal Memes},
 volume = {33},
 year = {2020}
}

@inproceedings{deng2009imagenet,
  title={Imagenet: A large-scale hierarchical image database},
  author={Deng, Jia and Dong, Wei and Socher, Richard and Li, Li-Jia and Li, Kai and Fei-Fei, Li},
  booktitle={2009 {IEEE} conference on computer vision and pattern recognition},
  pages={248--255},
  year={2009},
  organization={Ieee}
}

@inproceedings{
dao2023flashattention,
title={FlashAttention-2: Faster Attention with Better Parallelism and Work Partitioning},
author={Tri Dao},
booktitle={The Twelfth International Conference on Learning Representations},
year={2024},
url={https://openreview.net/forum?id=mZn2Xyh9Ec}
}

@inproceedings{ma2025tevatron,
  title={Tevatron 2.0: Unified document retrieval toolkit across scale, language, and modality},
  author={Ma, Xueguang and Gao, Luyu and Zhuang, Shengyao and Zhan, Jiaqi Samantha and Callan, Jamie and Lin, Jimmy},
  booktitle={Proceedings of the 48th International ACM SIGIR Conference on Research and Development in Information Retrieval},
  pages={4061--4065},
  year={2025}
}

@article{bai2025qwen2,
  title={Qwen2.5-vl technical report},
  author={Bai, Shuai and Chen, Keqin and Liu, Xuejing and Wang, Jialin and Ge, Wenbin and Song, Sibo and Dang, Kai and Wang, Peng and Wang, Shijie and Tang, Jun and Zhong, Humen and Zhu, Yuanzhi and Yang, Mingkun and Li, Zhaohai and Wan, Jianqiang and Wang, Pengfei and Ding, Wei and Fu, Zheren and Xu, Yiheng and Ye, Jiabo and Zhang, Xi and Xie, Tianbao and Cheng, Zesen and Zhang, Hang and Yang, Zhibo and Xu, Haiyang and Lin, Junyang and others},
  journal={arXiv preprint arXiv:2502.13923},
  year={2025}
}

@inproceedings{mustafa2022multimodal,
 author = {Mustafa, Basil and Riquelme, Carlos and Puigcerver, Joan and Jenatton, Rodolphe and Houlsby, Neil},
 booktitle = {Advances in Neural Information Processing Systems},
 pages = {9564--9576},
 title = {Multimodal Contrastive Learning with LIMoE: the Language-Image Mixture of Experts},
 volume = {35},
 year = {2022}
}

@InProceedings{Li_2024_CVPR,
    author    = {Li, Kunchang and Wang, Yali and He, Yinan and Li, Yizhuo and Wang, Yi and Liu, Yi and Wang, Zun and Xu, Jilan and Chen, Guo and Luo, Ping and Wang, Limin and Qiao, Yu},
    title     = {MVBench: A Comprehensive Multi-modal Video Understanding Benchmark},
    booktitle = {Proceedings of the IEEE/CVF Conference on Computer Vision and Pattern Recognition (CVPR)},
    month     = {June},
    year      = {2024},
    pages     = {22195--22206}
}

@InProceedings{Fu_2025_CVPR,
    author    = {Fu, Chaoyou and Dai, Yuhan and Luo, Yongdong and Li, Lei and Ren, Shuhuai and Zhang, Renrui and Wang, Zihan and Zhou, Chenyu and Shen, Yunhang and Zhang, Mengdan and Chen, Peixian and Li, Yanwei and Lin, Shaohui and Zhao, Sirui and Li, Ke and Xu, Tong and Zheng, Xiawu and Chen, Enhong and Shan, Caifeng and He, Ran and Sun, Xing},
    title     = {Video-MME: The First-Ever Comprehensive Evaluation Benchmark of Multi-modal LLMs in Video Analysis},
    booktitle = {Proceedings of the IEEE/CVF Conference on Computer Vision and Pattern Recognition (CVPR)},
    month     = {June},
    year      = {2025},
    pages     = {24108--24118}
}

@inproceedings{10.1609/aaai.v33i01.33019127,
  title={Activitynet-qa: A dataset for understanding complex web videos via question answering},
  author={Yu, Zhou and Xu, Dejing and Yu, Jun and Yu, Ting and Zhao, Zhou and Zhuang, Yueting and Tao, Dacheng},
  booktitle={Proceedings of the AAAI conference on artificial intelligence},
  pages={9127--9134},
  year={2019}
}

@inproceedings{NEURIPS2023_90ce332a,
 author = {Mangalam, Karttikeya and Akshulakov, Raiymbek and Malik, Jitendra},
 booktitle = {Advances in Neural Information Processing Systems},
 pages = {46212--46244},
 title = {EgoSchema: A Diagnostic Benchmark for Very Long-form Video Language Understanding},
 volume = {36},
 year = {2023}
}

@InProceedings{Xiao_2021_CVPR,
    author    = {Xiao, Junbin and Shang, Xindi and Yao, Angela and Chua, Tat-Seng},
    title     = {NExT-QA: Next Phase of Question-Answering to Explaining Temporal Actions},
    booktitle = {Proceedings of the IEEE/CVF Conference on Computer Vision and Pattern Recognition (CVPR)},
    month     = {June},
    year      = {2021},
    pages     = {9777--9786}
}

@inproceedings{
NEURIPS2025_281e0b91,
title={MomentSeeker: A Task-Oriented Benchmark For Long-Video Moment Retrieval},
author={Huaying Yuan and Jian Ni and Zheng Liu and Yueze Wang and Junjie Zhou and Zhengyang Liang and Bo Zhao and Zhao Cao and Ji-Rong Wen and Zhicheng Dou},
booktitle={The Thirty-ninth Annual Conference on Neural Information Processing Systems Datasets and Benchmarks Track},
year={2025},
url={https://openreview.net/forum?id=gKkA9Oc13m}
}

@inproceedings{gao2017tall,
  title={Tall: Temporal activity localization via language query},
  author={Gao, Jiyang and Sun, Chen and Yang, Zhenheng and Nevatia, Ram},
  booktitle={Proceedings of the IEEE international conference on computer vision},
  pages={5267--5275},
  year={2017}
}

@inproceedings{lei2021detecting,
 author = {Lei, Jie and Berg, Tamara L and Bansal, Mohit},
 booktitle = {Advances in Neural Information Processing Systems},
 pages = {11846--11858},
 title = {Detecting Moments and Highlights in Videos via Natural Language Queries},
 volume = {34},
 year = {2021}
}

@inproceedings{zhou2018towards,
  title={Towards automatic learning of procedures from web instructional videos},
  author={Zhou, Luowei and Xu, Chenliang and Corso, Jason},
  booktitle={Proceedings of the AAAI conference on artificial intelligence},
  year={2018}
}

@inproceedings{wang2019vatex,
  title={Vatex: A large-scale, high-quality multilingual dataset for video-and-language research},
  author={Wang, Xin and Wu, Jiawei and Chen, Junkun and Li, Lei and Wang, Yuan-Fang and Wang, William Yang},
  booktitle={Proceedings of the IEEE/CVF international conference on computer vision},
  pages={4581--4591},
  year={2019}
}

@inproceedings{wei2025boosting,
  title={Boosting multimodal learning via disentangled gradient learning},
  author={Wei, Shicai and Luo, Chunbo and Luo, Yang},
  booktitle={Proceedings of the IEEE/CVF International Conference on Computer Vision},
  pages={22879--22888},
  year={2025}
}

@inproceedings{
hao2026unix,
title={Uni-X: Mitigating Modality Conflict with a Two-End-Separated Architecture for Unified Multimodal Models},
author={Jitai Hao and Hao Liu and Xinyan Xiao and Qiang Huang and Jun Yu},
booktitle={The Fourteenth International Conference on Learning Representations},
year={2026},
url={https://openreview.net/forum?id=IJLIYpCkwz}
}

@inproceedings{
tang2026wave,
title={{WAVE}: Learning Unified \& Versatile Audio-Visual Embeddings with Multimodal {LLM}},
author={Changli Tang and Qinfan Xiao and Ke Mei and Tianyi Wang and Fengyun Rao and Chao Zhang},
booktitle={The Fourteenth International Conference on Learning Representations},
year={2026},
url={https://openreview.net/forum?id=MiV3WXDYJb}
}

@article{xu2025omni,
  title={Omni-Embed-Nemotron: A Unified Multimodal Retrieval Model for Text, Image, Audio, and Video},
  author={Xu, Mengyao and Zhou, Wenfei and Babakhin, Yauhen and Moreira, Gabriel and Ak, Ronay and Osmulski, Radek and Liu, Bo and Oldridge, Even and Schifferer, Benedikt},
  journal={arXiv preprint arXiv:2510.03458},
  year={2025}
}

@article{xu2026scaling,
  title={Scaling Audio-Text Retrieval with Multimodal Large Language Models},
  author={Xu, Jilan and Thom{\'e}, Carl and Horak, Danijela and Xie, Weidi and Zisserman, Andrew},
  journal={arXiv preprint arXiv:2602.18010},
  year={2026}
}

@article{jiang2024e5,
  title={E5-v: Universal embeddings with multimodal large language models},
  author={Jiang, Ting and Song, Minghui and Zhang, Zihan and Huang, Haizhen and Deng, Weiwei and Sun, Feng and Zhang, Qi and Wang, Deqing and Zhuang, Fuzhen},
  journal={arXiv preprint arXiv:2407.12580},
  year={2024}
}

@article{guo2025towards,
  title={Towards Universal Video Retrieval: Generalizing Video Embedding via Synthesized Multimodal Pyramid Curriculum},
  author={Guo, Zhuoning and Li, Mingxin and Zhang, Yanzhao and Long, Dingkun and Xie, Pengjun and Chu, Xiaowen},
  journal={arXiv preprint arXiv:2510.27571},
  year={2025}
}

@inproceedings{
hu2022lora,
title={Lo{RA}: Low-Rank Adaptation of Large Language Models},
author={Edward J Hu and Yelong Shen and Phillip Wallis and Zeyuan Allen-Zhu and Yuanzhi Li and Shean Wang and Lu Wang and Weizhu Chen},
booktitle={International Conference on Learning Representations},
year={2022},
url={https://openreview.net/forum?id=nZeVKeeFYf9}
}

@inproceedings{
xiao2026scaling,
title={Scaling Language-centric Omnimodal Representation Learning},
author={Chenghao Xiao and Hou Pong Chan and Hao Zhang and Weiwen Xu and Mahani Aljunied and Yu Rong},
booktitle={The Thirty-ninth Annual Conference on Neural Information Processing Systems},
year={2025},
url={https://openreview.net/forum?id=eA3Aum0rpA}
}

@article{xu2025qwen25,
  title={Qwen2.5-Omni Technical Report},
  author={Xu, Jin and Guo, Zhifang and He, Jinzheng and Hu, Hangrui and He, Ting and Bai, Shuai and Chen, Keqin and Wang, Jialin and Fan, Yang and Dang, Kai and Zhang, Bin and Wang, Xiong and Chu, Yunfei and Lin, Junyang},
  journal={arXiv preprint arXiv:2503.20215},
  year={2025}
}

@article{
meng2026vlmvecv,
title={{VLM}2Vec-V2: Advancing Multimodal Embedding for Videos, Images, and Visual Documents},
author={Rui Meng and Ziyan Jiang and Ye Liu and Mingyi Su and Xinyi Yang and Yuepeng Fu and Can Qin and Raghuveer Thirukovalluru and Xuan Zhang and Zeyuan Chen and Ran Xu and Caiming Xiong and Yingbo Zhou and Wenhu Chen and Semih Yavuz},
journal={Transactions on Machine Learning Research},
issn={2835-8856},
year={2026},
url={https://openreview.net/forum?id=TpU38jbKIJ}
}

@article{wu2025mca,
  title={MCA: Modality Composition Awareness for Robust Composed Multimodal Retrieval},
  author={Wu, Qiyu and Cui, Shuyang and Hayakawa, Satoshi and Wang, Wei-Yao and Wakaki, Hiromi and Mitsufuji, Yuki},
  journal={arXiv preprint arXiv:2510.15543},
  year={2025}
}

@string{acl = {Association for Computational Linguistics}}

@string{anth = {https://aclanthology.org/}}

@inproceedings{chen-etal-2026-e5,title = "e5-omni: Explicit Cross-modal Alignment for Omni-modal Embeddings",author = "Chen, Haonan and Gao, Sicheng and Timofte, Radu and Sakai, Tetsuya and Dou, Zhicheng",editor = "Liakata, Maria and Moreira, Viviane P. and Zhang, Jiajun and Jurgens, David",booktitle = "Findings of the {A}ssociation for {C}omputational {L}inguistics: {ACL} 2026",month = jul,year = "2026",address = "San Diego, California, United States",publisher = acl,url = anth # {2026.findings-acl.970/},doi = "10.18653/v1/2026.findings-acl.970",pages = "19430--19443",ISBN = "979-8-89176-395-1"}

@inproceedings{zhang-etal-2025-direct,title = "Direct Preference Optimization of Video Large Multimodal Models from Language Model Reward",author = "Zhang, Ruohong and Gui, Liangke and Sun, Zhiqing and Feng, Yihao and Xu, Keyang and Zhang, Yuanhan and Fu, Di and Li, Chunyuan and Hauptmann, Alexander G and Bisk, Yonatan and Yang, Yiming",editor = "Chiruzzo, Luis and Ritter, Alan and Wang, Lu",booktitle = "Proceedings of the 2025 Conference of the Nations of the Americas Chapter of the Association for Computational Linguistics: Human Language Technologies (Volume 1: Long Papers)",month = apr,year = "2025",address = "Albuquerque, New Mexico",publisher = acl,url = anth # {2025.naacl-long.30/},doi = "10.18653/v1/2025.naacl-long.30",pages = "694--717",ISBN = "979-8-89176-189-6"}

@inproceedings{melechovsky-etal-2024-mustango,title = "Mustango: Toward Controllable Text-to-Music Generation",author = "Melechovsky, Jan and Guo, Zixun and Ghosal, Deepanway and Majumder, Navonil and Herremans, Dorien and Poria, Soujanya",editor = "Duh, Kevin and Gomez, Helena and Bethard, Steven",booktitle = "Proceedings of the 2024 Conference of the North American Chapter of the Association for Computational Linguistics: Human Language Technologies (Volume 1: Long Papers)",month = jun,year = "2024",address = "Mexico City, Mexico",publisher = acl,url = anth # {2024.naacl-long.459/},doi = "10.18653/v1/2024.naacl-long.459",pages = "8293--8316"}

@inproceedings{deng-etal-2024-musilingo,title = "{M}usi{L}ingo: Bridging Music and Text with Pre-trained Language Models for Music Captioning and Query Response",author = "Deng, Zihao and Ma, Yinghao and Liu, Yudong and Guo, Rongchen and Zhang, Ge and Chen, Wenhu and Huang, Wenhao and Benetos, Emmanouil",editor = "Duh, Kevin and Gomez, Helena and Bethard, Steven",booktitle = "Findings of the Association for Computational Linguistics: NAACL 2024",month = jun,year = "2024",address = "Mexico City, Mexico",publisher = acl,url = anth # {2024.findings-naacl.231/},doi = "10.18653/v1/2024.findings-naacl.231",pages = "3643--3655"}

@inproceedings{ma-etal-2024-modula,title = "{M}o{DULA}: Mixture of Domain-Specific and Universal {L}o{RA} for Multi-Task Learning",author = "Ma, Yufei and Liang, Zihan and Dai, Huangyu and Chen, Ben and Gao, Dehong and Ran, Zhuoran and Zihan, Wang and Jin, Linbo and Jiang, Wen and Zhang, Guannan and Cai, Xiaoyan and Yang, Libin",editor = "Al-Onaizan, Yaser and Bansal, Mohit and Chen, Yun-Nung",booktitle = "Proceedings of the 2024 Conference on Empirical Methods in Natural Language Processing",month = nov,year = "2024",address = "Miami, Florida, USA",publisher = acl,url = anth # {2024.emnlp-main.161/},doi = "10.18653/v1/2024.emnlp-main.161",pages = "2758--2770"}

@inproceedings{ma-etal-2024-unifying,title = "Unifying Multimodal Retrieval via Document Screenshot Embedding",author = "Ma, Xueguang and Lin, Sheng-Chieh and Li, Minghan and Chen, Wenhu and Lin, Jimmy",editor = "Al-Onaizan, Yaser and Bansal, Mohit and Chen, Yun-Nung",booktitle = "Proceedings of the 2024 Conference on Empirical Methods in Natural Language Processing",month = nov,year = "2024",address = "Miami, Florida, USA",publisher = acl,url = anth # {2024.emnlp-main.373/},doi = "10.18653/v1/2024.emnlp-main.373",pages = "6492--6505"}

@inproceedings{liu-etal-2023-edis,title = "{EDIS}: Entity-Driven Image Search over Multimodal Web Content",author = "Liu, Siqi and Feng, Weixi and Fu, Tsu-Jui and Chen, Wenhu and Wang, William",editor = "Bouamor, Houda and Pino, Juan and Bali, Kalika",booktitle = "Proceedings of the 2023 Conference on Empirical Methods in Natural Language Processing",month = dec,year = "2023",address = "Singapore",publisher = acl,url = anth # {2023.emnlp-main.297/},doi = "10.18653/v1/2023.emnlp-main.297",pages = "4877--4894"}

@inproceedings{wang-etal-2022-n24news,title = "{N}24{N}ews: A New Dataset for Multimodal News Classification",author = "Wang, Zhen and Shan, Xu and Zhang, Xiangxie and Yang, Jie",editor = "Calzolari, Nicoletta and B{\'e}chet, Fr{\'e}d{\'e}ric and Blache, Philippe and Choukri, Khalid and Cieri, Christopher and Declerck, Thierry and Goggi, Sara and Isahara, Hitoshi and Maegaard, Bente and Mariani, Joseph and Mazo, H{\'e}l{\`e}ne and Odijk, Jan and Piperidis, Stelios",booktitle = "Proceedings of the Thirteenth Language Resources and Evaluation Conference",month = jun,year = "2022",address = "Marseille, France",publisher = "European Language Resources Association",url = anth # {2022.lrec-1.729/},pages = "6768--6775"}

@inproceedings{masry-etal-2022-chartqa,title = "{C}hart{QA}: A Benchmark for Question Answering about Charts with Visual and Logical Reasoning",author = "Masry, Ahmed and Long, Do Xuan and Tan, Jia Qing and Joty, Shafiq and Hoque, Enamul",editor = "Muresan, Smaranda and Nakov, Preslav and Villavicencio, Aline",booktitle = "Findings of the Association for Computational Linguistics: ACL 2022",month = may,year = "2022",address = "Dublin, Ireland",publisher = acl,url = anth # {2022.findings-acl.177/},doi = "10.18653/v1/2022.findings-acl.177",pages = "2263--2279"}

@inproceedings{gao-etal-2021-scaling,title = "Scaling Deep Contrastive Learning Batch Size under Memory Limited Setup",author = "Gao, Luyu and Zhang, Yunyi and Han, Jiawei and Callan, Jamie",editor = "Rogers, Anna and Calixto, Iacer and Vuli{\'c}, Ivan and Saphra, Naomi and Kassner, Nora and Camburu, Oana-Maria and Bansal, Trapit and Shwartz, Vered",booktitle = "Proceedings of the 6th Workshop on Representation Learning for NLP (RepL4NLP-2021)",month = aug,year = "2021",address = "Online",publisher = acl,url = anth # {2021.repl4nlp-1.31/},doi = "10.18653/v1/2021.repl4nlp-1.31",pages = "316--321"}

@inproceedings{liu-etal-2021-visual,title = "Visual News: Benchmark and Challenges in News Image Captioning",author = "Liu, Fuxiao and Wang, Yinghan and Wang, Tianlu and Ordonez, Vicente",editor = "Moens, Marie-Francine and Huang, Xuanjing and Specia, Lucia and Yih, Scott Wen-tau",booktitle = "Proceedings of the 2021 Conference on Empirical Methods in Natural Language Processing",month = nov,year = "2021",address = "Online and Punta Cana, Dominican Republic",publisher = acl,url = anth # {2021.emnlp-main.542/},doi = "10.18653/v1/2021.emnlp-main.542",pages = "6761--6771"}

@inproceedings{kim-etal-2019-audiocaps,title = "{A}udio{C}aps: Generating Captions for Audios in The Wild",author = "Kim, Chris Dongjoo and Kim, Byeongchang and Lee, Hyunmin and Kim, Gunhee",editor = "Burstein, Jill and Doran, Christy and Solorio, Thamar",booktitle = "Proceedings of the 2019 Conference of the North {A}merican Chapter of the Association for Computational Linguistics: Human Language Technologies, Volume 1 (Long and Short Papers)",month = jun,year = "2019",address = "Minneapolis, Minnesota",publisher = acl,url = anth # {N19-1011/},doi = "10.18653/v1/N19-1011",pages = "119--132"}

@inproceedings{kazemzadeh-etal-2014-referitgame,title = "{R}efer{I}t{G}ame: Referring to Objects in Photographs of Natural Scenes",author = "Kazemzadeh, Sahar and Ordonez, Vicente and Matten, Mark and Berg, Tamara",editor = "Moschitti, Alessandro and Pang, Bo and Daelemans, Walter",booktitle = "Proceedings of the 2014 Conference on Empirical Methods in Natural Language Processing ({EMNLP})",month = oct,year = "2014",address = "Doha, Qatar",publisher = acl,url = anth # {D14-1086/},doi = "10.3115/v1/D14-1086",pages = "787--798"}

@inproceedings{chen-dolan-2011-collecting,title = "Collecting Highly Parallel Data for Paraphrase Evaluation",author = "Chen, David and Dolan, William",editor = "Lin, Dekang and Matsumoto, Yuji and Mihalcea, Rada",booktitle = "Proceedings of the 49th Annual Meeting of the Association for Computational Linguistics: Human Language Technologies",month = jun,year = "2011",address = "Portland, Oregon, USA",publisher = acl,url = anth # {P11-1020/},pages = "190--200"}

\clearpage
\appendix
\addtocontents{toc}{\protect\setcounter{tocdepth}{2}}
\tableofcontents

\section{Additional Details}
\subsection{Training Datasets}
\label{sec:train_data}
\begin{figure}[t]
  \centering
   \includegraphics[width=0.99\columnwidth]{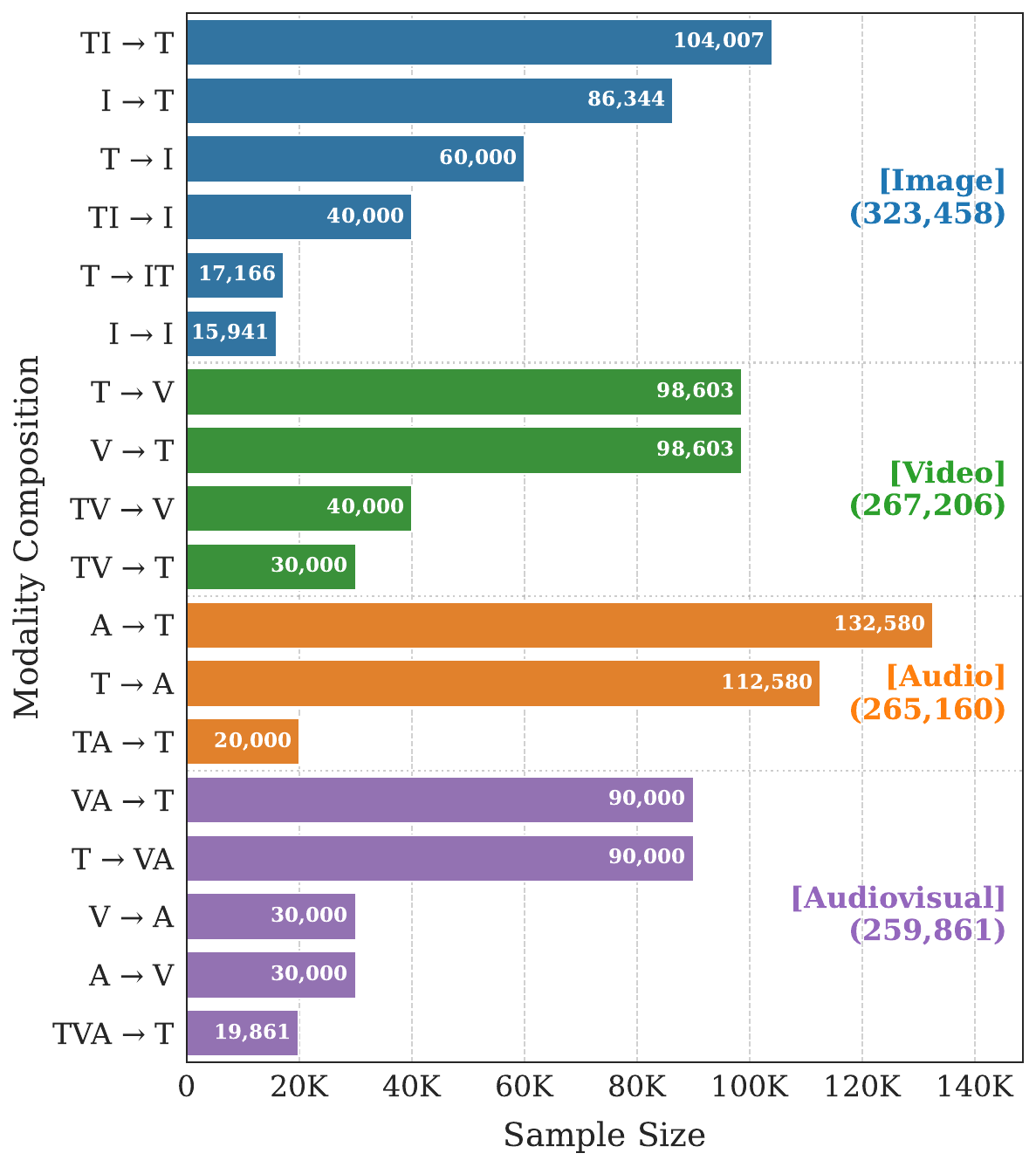}
    \caption{Distribution of sample counts for each query-to-target modality composition in our training mixture. For each composition, `T', `I', `V', and `A' correspond to text, image, video, and audio modality, respectively.
    We also specify the total sample size per modality group.
    }
   \label{fig:training_stat}
\end{figure}

To facilitate omnimodal learning, we construct a comprehensive training mixture categorized into four modality groups: image, video, audio, and audiovisual. Beyond single modalities, this mixture features composed modality queries and/or targets, allowing the model to learn versatile alignment spaces. The distribution and coverage of sample counts for each query-to-target modality composition are visualized in \cref{fig:training_stat}, while detailed dataset-level specifications for each modality group are provided in~\cref{tab:omni_image,tab:omni_video,tab:omni_audio,tab:omni_audiovisual}, respectively. 
For each dataset, if the original dataset size exceeded our specified target size, we randomly sub-sampled it to match that specified size. 

As shown in~\cref{fig:teaser}, while the modality-specific models were trained independently only on the datasets corresponding to their respective modality groups, our \baseline\ and \mymethod\ models were jointly trained on the entire mixture encompassing all modalities simultaneously.

\subsection{Evaluation Datasets}
\label{sec:eval_data}
To construct our evaluation suite across four distinct modalities: image, video, audio, and audiovisual, we leverage and extend existing evaluation frameworks. Specifically, for the image and video tasks, we utilize the evaluation suites from MMEB-v1~\cite{jiang2025vlmvec} and MMEB-v2~\cite{meng2026vlmvecv}, respectively. While MMEB-v2 natively includes text-to-video retrieval, it lacks the corresponding video-to-text direction. Therefore, we integrated video-to-text retrieval to ensure a bidirectional evaluation. Furthermore, we expand the benchmark's scope by incorporating additional audio and audiovisual tasks.

The detailed statistics and configurations for each modality are provided from Tab.~\ref{tab:supple_eval_image} to Tab.~\ref{tab:supple_eval_audiovisual}, each specifying the task type, query-to-target composition, and data sizes. In these tables, the corpus size is defined differently based on the task type: it denotes the number of classes for classification (CLS) tasks, the number of choices for question answering (QA) tasks, and the actual database size for retrieval (RET) tasks. Our data collection includes a wide variety of common-domain settings for each modality, such as integrating both generic audio and music domains. English is the primary language across all multi-modal tasks.

\section{Detailed Experimental Settings}
\subsection{Training Details \& Hyperparameters}
\label{sec:supple_implementation}
\noindent \textbf{Model Backbone.} 
We select the pretrained Qwen-2.5-Omni~\cite{xu2025qwen25} as the backbone. During training, the visual and audio encoders are frozen, and LoRA parameters are inserted into the LLM layers. The LoRA target modules are \texttt{q\_proj}, \texttt{k\_proj}, \texttt{v\_proj}, \texttt{o\_proj}, \texttt{down\_proj}, \texttt{up\_proj}, and \texttt{gate\_proj}, which are applied to each of the 36 LLM layers of the Qwen-2.5-Omni backbone.

\noindent \textbf{Input Processing.} 
For the input processing, we adopt the native configuration of Qwen-2.5-Omni~\cite{xu2025qwen25} unless specified otherwise. Specifically, images and videos are processed using the native resolution scaling strategy of Qwen-2.5-VL~\cite{bai2025qwen2}. The minimum and maximum image resolutions are set to 50,176 pixels ($64 \times 28 \times 28$) and 784,000 pixels ($1000 \times 28 \times 28$), respectively. For video inputs, videos are uniformly sampled into 8 frames, and the per-frame resolution is bounded between a minimum of 100,352 pixels ($128 \times 28 \times 28$) and a maximum of 156,800 pixels ($200 \times 28 \times 28$). Audio signals are sampled at 16,000 Hz. The textual inputs are truncated to a maximum of 512 text tokens (without chat templates), and consistent with prior approaches~\cite{meng2026vlmvecv,chen-etal-2026-e5}, left-padding is applied during batching.

\noindent \textbf{LoRA Configurations.} 
For the modality-specific models and the standard LoRA baseline \baseline\ mentioned in~\cref{fig:teaser}, the LoRA rank ($r$) is 16. For \mymethod\ with OME-LoRA, the rank of the shared adaptation path is 8, and each modality-expert rank is 4, making the total effective rank 24. Across all configurations, the LoRA scaling factor $\alpha$ is set to two times the rank. LoRA dropout is applied with probability 0.1. 
The detailed parameter counts for both the \baseline\ baseline and \mymethod\ variants under these LoRA configurations are summarized in~\cref{tab:model_parameters_comparison}. Notably, the modality router introduces minimal parameter overhead, less than 8\% of the total trainable parameters.

\begin{table}[t]
\centering
\resizebox{.99\columnwidth}{!}{%
    \begin{tabular}{lcccc}
    \toprule
    \multirow{2}{*}{} & \multicolumn{2}{c}{\baseline} & \multicolumn{2}{c}{\mymethod} \\
    \cmidrule(lr){2-3} \cmidrule(lr){4-5}
     & {3B} & {7B} & {3B} & {7B} \\
    \midrule
    Backbone & \multicolumn{4}{c}{Qwen-2.5-Omni} \\
    \midrule
    LoRA rank (Shared) &  \multicolumn{2}{c}{16} & \multicolumn{2}{c}{8}  \\
    LoRA rank (Expert) &  \multicolumn{2}{c}{0} & \multicolumn{2}{c}{$4 \times 4$} \\
    \cmidrule(lr){1-5}
    LoRA rank (Total) & \multicolumn{2}{c}{16} & \multicolumn{2}{c}{24} \\
    \midrule
    LoRA params.    & 29.9 M   & 40.4 M   & 44.9 M   & 60.6 M   \\
    Router params.  & 0.0 M    & 0.0 M    & 3.4 M    & 4.5 M    \\
    \cmidrule(lr){1-5}
    Total trainable params.   & 29.9 M   & 40.4 M   & 48.3 M   & 65.1 M   \\
    \bottomrule
    \end{tabular}
}
\caption{Summary of LoRA configuration and trainable parameters for \baseline\ and \mymethod\ model variants.}
\label{tab:model_parameters_comparison}
\end{table}

\noindent \textbf{Optimization.} 
All models are trained for one epoch using the Hugging Face Trainer with FlashAttention-2~\cite{dao2023flashattention} in BFloat16 mixed-precision. We use the AdamW optimizer with $\beta_1 = 0.9$, $\beta_2 = 0.999$, and $\epsilon = 10^{-8}$. We employ a cosine learning rate schedule with a peak learning rate of $10^{-4}$, a warmup ratio of 0.04, a weight decay of 0.01, and a maximum gradient norm of 1.0. The global batch size is set to 768. We employ GradCache~\cite{gao-etal-2021-scaling} to increase the global batch size, where the chunk size is set to 1 for the 3B models and 4 for the 7B models. We do not employ hard negative examples. The 3B models are trained using 8$\times$ NVIDIA A100 (40GB) GPUs, which takes approximately 5 days for \texttt{Syn-Omni}, and the 7B models are trained using 8$\times$ NVIDIA H200 (141GB) GPUs, taking approximately 40 hours for \texttt{Syn-Omni}.

\noindent \textbf{Training Objective.} 
The baseline models and modality-specific models are optimized using the contrastive loss. \texttt{Syn-Omni} is optimized using the contrastive loss combined with the orthogonal penalty and the synergy margin loss as proposed in the main paper. The contrastive loss uses a temperature parameter $\tau$ of 0.02. For the orthogonal penalty, the loss weight is set to 0.2. The Progressive Synergy Routing (PSR) schedule uses warmup and transition ratios of $t_{\text{warm}} = 0.1$ and $t_{\text{blend}} = 0.5$. For the synergy margin loss, the margin $\gamma$ is 0.2 and the loss weight is set to 0.2.

\subsection{Evaluation Pipeline} 
For a fair evaluation, previous methods are evaluated under the unified setting using the official checkpoints. We extend the VLM2Vec-V2 codebase used for evaluation~\cite{meng2026vlmvecv}, which enables the integrated evaluation of diverse modalities and task types while facilitating seamless modality expansion. While this evaluation framework originally supports image and video inputs, we extend it to encompass audio and audiovisual modalities also, thereby enabling an any-to-any query-to-target retrieval setting.

The input processing for text, image, video, and audio modalities remains identical to those used during the training phase. For inference, standard LoRA adapter parameters are merged into the base pretrained weights, and such LoRA layers are not utilized any further. In contrast, for \texttt{Syn-Omni}, only the shared adaptation path is merged into the backbone weights and removed, whereas the expert paths are explicitly maintained to support dynamic routing during inference, where the soft routing weights generated by the router are utilized.

\begin{figure*}[t]
  \centering
   \includegraphics[width=0.95\linewidth]{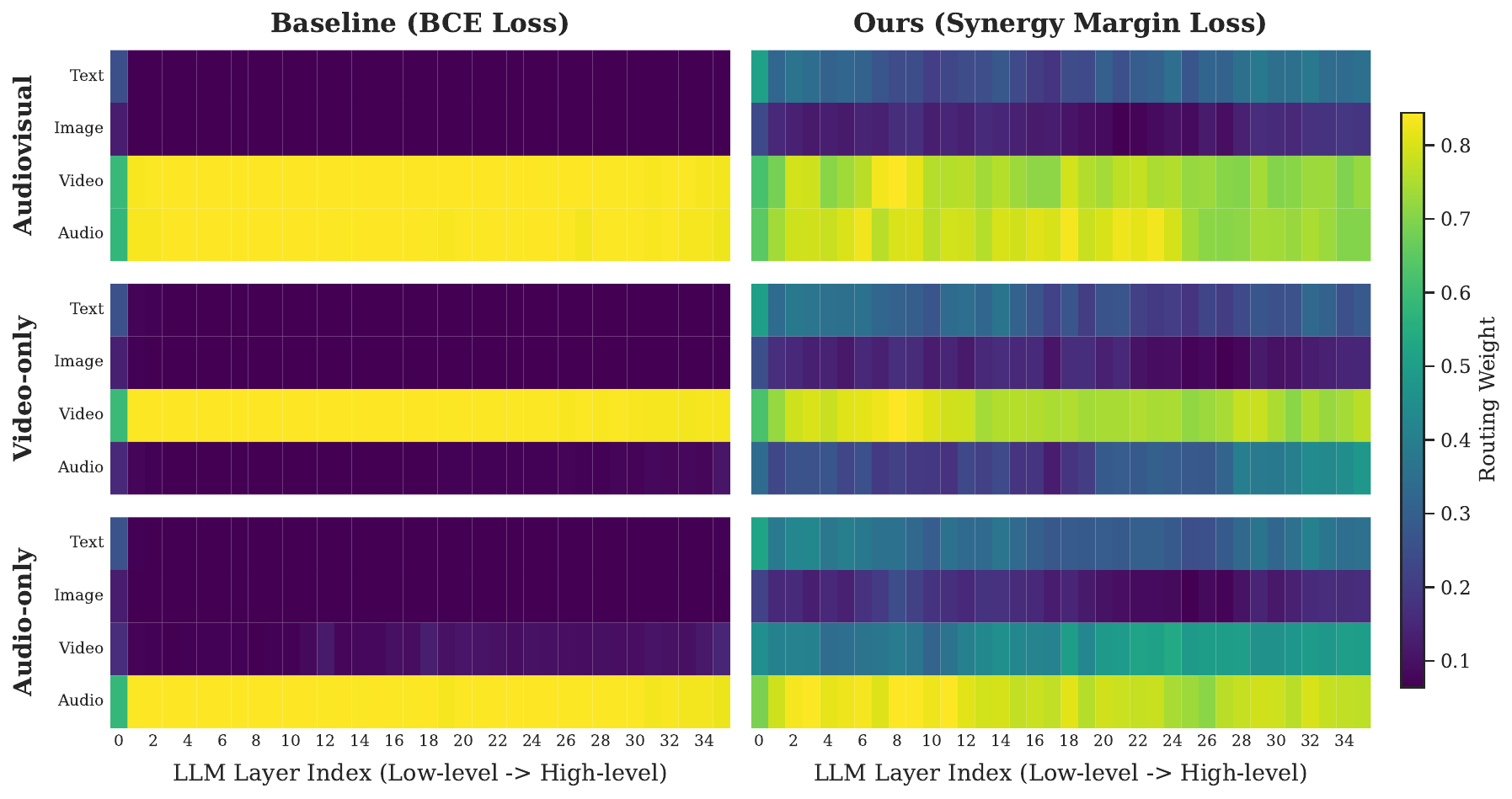}
    \caption{Visualization of layer-wise soft routing weights under different input modalities. The baseline (BCE loss) exhibits a collapsed hard routing behavior with nearly binary activations, whereas ours (synergy margin loss) dynamically allocates routing weights even to absent-modality experts to encourage cross-modal synergy.}
   \label{fig:supple_heatmap}
\end{figure*}

\section{Detailed Results}

\subsection{Routing Behavior across Layers}
Extending the analysis in \cref{fig:progress_routing}, we investigate how different router loss formulations affect routing dynamics at the layer level. While the main paper tracks routing weights across training steps, this section visualizes the layer-wise soft routing activations of the converged model using audiovisual, video-only, and audio-only inputs from VALOR-32k~\cite{10721284}. Under this configuration, \cref{fig:supple_heatmap} compares the binary cross-entropy (BCE) loss against our synergy margin loss across the entire layer hierarchy.

As shown in the figure, BCE loss forces the router into strict modality classification based on the input type. Except for the first layer, this results in routing activations approaching 1 for present modalities and 0 for absent ones, collapsing into a deterministic hard routing behavior despite using a soft routing formulation. In contrast, our synergy margin loss maintains dominant activations for the input modalities while preserving meaningful activations even for absent modalities across all layers. This collaborative activation behavior across the layer hierarchy successfully facilitates cross-modal synergy, aligning with our architectural objectives.

\subsection{Modality-Specific Models}
\label{sec:modality_specific}
We present the extended evaluation of the modality-specific models introduced in \cref{fig:teaser}, reporting their performance across all modalities in \cref{tab:modality_specific}. Models trained exclusively on a single modality demonstrate strong performance within their respective modality. For instance, the audio-specific model performs well on audio tasks. A notable exception is observed in the audiovisual-specific model. This indicates that exposure to individual image, video, and audio data within a unified training scheme is helpful for audiovisual tasks. 

\begin{table}[t]
\centering
\resizebox{.99\linewidth}{!}{%
\begin{tabular}{lccccc}
\toprule
      &   I &   V &   A &   AV &   All \\
\midrule
 I-specific      &  68.0 &  43.0 &  35.4 &        33.3 &           44.9 \\
 V-specific      &  48.1 &  41.1 &  34.9 &        35.9 &           40.0 \\
 A-specific      &  40.3 &  36.5 &  43.5 &        30.9 &           37.8 \\
 AV-specific     &  38.0 &  32.4 &  34.6 &        37.1 &           35.5 \\
\cmidrule(lr){1-6}
 \baseline\ (3B)       &  66.8 &  40.4 &  43.1 &        40.4 &           47.7 \\
 \mymethod\ (3B)       &  68.1 &  40.5 &  45.6 &        43.4 &           49.4\\
\bottomrule
\end{tabular}
}
\caption{Comprehensive evaluation of modality-specific models on all modalities. The results demonstrate intrinsic cross-modal transferability and highlight the necessity of unified multimodal training for cross-modal synergy. Here, I, V, A, and AV denote image, video, audio, and audiovisual modalities, respectively.}
\label{tab:modality_specific}
\end{table}

\begin{table}[t]
\centering
\resizebox{.99\columnwidth}{!}{%
    \begin{tabular}{lcc}
    \toprule
    {Model} & {Latency (ms)} & {Peak GPU Mem. (MB)} \\
    \midrule
    \baseline\ (3B) & 300.3 $\pm$ 25 & 9612 \\
    \mymethod\ (3B) & 354.6 $\pm$ 20 & 9676 \\
    \midrule
    \baseline\ (7B) & 377.1 $\pm$ 26 & 17672 \\
    \mymethod\ (7B) & 443.5 $\pm$ 30 & 17765 \\
    \bottomrule
    \end{tabular}
}
\caption{Runtime analysis on \baseline\ and our \mymethod\ model variants. We report mean latency (ms) and peak GPU memory usage (MB), testing on VALOR-32k audiovisual input samples.}
\label{tab:runtime_analysis}
\end{table}

\begin{table*}[t]
\centering
\resizebox{.65\linewidth}{!}{%
    \begin{tabular}{lcc}
        \toprule
        {Dataset} & {Composition} & {Size} \\
        \midrule
        ImageNet 1K~\cite{deng2009imagenet} & I $\rightarrow$ T & 15,000 \\
        N24News~\cite{wang-etal-2022-n24news} & TI $\rightarrow$ T & 15,000 \\
        HatefulMemes~\cite{kiela2020hateful} & I $\rightarrow$ T & 8,500 \\
        VOC2007~\cite{everingham2015pascal} & I $\rightarrow$ T & 7,844 \\
        SUN397~\cite{xiao2010sun} & I $\rightarrow$ T & 15,000 \\
        OK-VQA~\cite{marino2019ok} & TI $\rightarrow$ T & 9,007 \\
        A-OKVQA~\cite{schwenk2022okvqa} & TI $\rightarrow$ T & 15,000 \\
        DocVQA~\cite{mathew2021docvqa} & TI $\rightarrow$ T & 15,000 \\
        InfographicsVQA~\cite{mathew2022infographicvqa} & TI $\rightarrow$ T & 15,000 \\
        ChartQA~\cite{masry-etal-2022-chartqa} & TI $\rightarrow$ T & 15,000 \\
        Visual7W~\cite{zhu2016visual7w} & TI $\rightarrow$ T & 20,000 \\
        VisDial~\cite{das2017visual} & T $\rightarrow$ I & 20,000 \\
        CIRR~\cite{liu2021image} & TI $\rightarrow$ I & 20,000 \\ 
        \cmidrule(lr){1-3}
        \multirow{2}{*}{VisualNews~\cite{liu-etal-2021-visual}} & T $\rightarrow$ I & 20,000 \\
                                    & I $\rightarrow$ T & 20,000 \\ 
        \cmidrule(lr){1-3}
        \multirow{2}{*}{MSCOCO-Retrieval~\cite{lin2014microsoft}}     & T $\rightarrow$ I & 20,000 \\
                                    & I $\rightarrow$ T & 20,000 \\
        \cmidrule(lr){1-3}
        MSCOCO-Grounding~\cite{lin2014microsoft} & TI $\rightarrow$ I & 20,000 \\
        \cmidrule(lr){1-3}
        NIGHTS~\cite{fu2023dreamsim} & I $\rightarrow$ I & 15,941 \\
        WebQA~\cite{chang2022webqa} & T $\rightarrow$ IT & 17,166 \\
        \midrule
        \textbf{Sum} & & \textbf{323,458} \\
        \bottomrule
    \end{tabular}
}
\caption{An overview of image-text datasets in the omnimodal training dataset. Each dataset is sourced from the MMEB-v1~\cite{jiang2025vlmvec}, where either the full original set or a randomly sampled subset was utilized.}
\label{tab:omni_image}
\end{table*}

\begin{table*}[t]
\centering
\resizebox{.65\linewidth}{!}{%
    \begin{tabular}{lcc}
        \toprule
        {Dataset} & {Composition} & {Size} \\ 
        \midrule
        \multirow{2}{*}{LLaVAHound Retrieval~\cite{zhang-etal-2025-direct}} & T $\rightarrow$ V & 30,000 \\
                                    & V $\rightarrow$ T & 30,000 \\
        \cmidrule(lr){1-3}
        LLaVAHound QA~\cite{zhang-etal-2025-direct} & TV $\rightarrow$ T & 30,000 \\ \cmidrule(lr){1-3}
        \multirow{2}{*}{PE-Video~\cite{bolya2026perception}}        & V $\rightarrow$ T & 40,000 \\
                                    & T $\rightarrow$ V & 40,000 \\ \cmidrule(lr){1-3}
        \multirow{2}{*}{MSVD~\cite{chen-dolan-2011-collecting}}       & V $\rightarrow$ T & 1,200 \\
                                    & T $\rightarrow$ V & 1,200 \\ \cmidrule(lr){1-3}
        \multirow{2}{*}{MSR-VTT~\cite{xu2016msr}}     & V $\rightarrow$ T & 9,000 \\
                                    & T $\rightarrow$ V & 9,000 \\ \cmidrule(lr){1-3}
        \multirow{2}{*}{ActivityNet~\cite{krishna2017dense}} & V $\rightarrow$ T & 10,008 \\
                                    & T $\rightarrow$ V & 10,008 \\ \cmidrule(lr){1-3}
        \multirow{2}{*}{DiDeMo~\cite{anne2017localizing}}     & V $\rightarrow$ T & 8,395 \\
                                    & T $\rightarrow$ V & 8,395 \\ \cmidrule(lr){1-3}
        FineCVR~\cite{wu2025learning}                     & TV $\rightarrow$ V & 40,000 \\
        \midrule
        \textbf{Sum} & & \textbf{267,206} \\ 
        \bottomrule
    \end{tabular}
}
\caption{An overview of video-centric datasets included in the omnimodal training dataset.}
\label{tab:omni_video}
\end{table*}

\begin{table*}[t]
\centering
\resizebox{.70\linewidth}{!}{%
    \begin{tabular}{lcc}
        \toprule
        {Dataset} & {Composition} & {Size} \\ 
        \midrule
        \multirow{2}{*}{AudioCaps~\cite{kim-etal-2019-audiocaps}}   & A $\rightarrow$ T & 45,000 \\
                                     & T $\rightarrow$ A & 45,000 \\ \cmidrule(lr){1-3}
        \multirow{2}{*}{WavCaps~\cite{mei2024wavcaps}}     & A $\rightarrow$ T & 45,000 \\
                                     & T $\rightarrow$ A & 45,000 \\ \cmidrule(lr){1-3}
        AudioSet-SL~\cite{gemmeke2017audio}             & A $\rightarrow$ T & 20,000 \\ \cmidrule(lr){1-3}
        \multirow{2}{*}{MusicCaps~\cite{agostinelli2023musiclm}}   & A $\rightarrow$ T & 2,580 \\
                                     & T $\rightarrow$ A & 2,580 \\ \cmidrule(lr){1-3}
        \multirow{2}{*}{MusicBench~\cite{melechovsky-etal-2024-mustango}}  & A $\rightarrow$ T & 20,000 \\
                                     & T $\rightarrow$ A & 20,000 \\ \cmidrule(lr){1-3}
        MusicInstruct~\cite{deng-etal-2024-musilingo}                & TA $\rightarrow$ T & 20,000 \\
        \midrule
        \textbf{Sum} & & \textbf{265,160} \\ 
        \bottomrule
    \end{tabular}
}
\caption{An overview of audio-centric datasets included in the omnimodal training dataset.}
\label{tab:omni_audio}
\end{table*}

\begin{table*}[t]
\centering
\resizebox{.65\linewidth}{!}{%
    \begin{tabular}{lcc}
        \toprule
        {Dataset} & {Composition} & {Size} \\ 
        \midrule
        \multirow{2}{*}{AudioSet~\cite{gemmeke2017audio}}     & V $\rightarrow$ A & 30,000 \\
                                      & A $\rightarrow$ V & 30,000 \\ \cmidrule(lr){1-3}
        \multirow{2}{*}{VAST~\cite{chen2023vast}}         & VA $\rightarrow$ T & 30,000 \\
                                      & T $\rightarrow$ VA & 30,000 \\ \cmidrule(lr){1-3}
        \multirow{2}{*}{InternVideo2~\cite{wang2024internvideo2}} & VA $\rightarrow$ T & 30,000 \\
                                      & T $\rightarrow$ VA & 30,000 \\ \cmidrule(lr){1-3}
        \multirow{2}{*}{Shot2Story~\cite{han2025shot2story}}   & VA $\rightarrow$ T & 30,000 \\
                                      & T $\rightarrow$ VA & 30,000 \\ \cmidrule(lr){1-3}
        AVInstruct~\cite{ye2024cat}                    & TVA $\rightarrow$ T & 19,861 \\
        \midrule
        \textbf{Sum} & & \textbf{259,861} \\ 
        \bottomrule
    \end{tabular}
}
\caption{An overview of audiovisual-centric datasets included in the omnimodal training dataset.}
\label{tab:omni_audiovisual}
\end{table*}

\begin{table*}[t]
\centering
\resizebox{.95\linewidth}{!}{%
    \begin{tabular}{clccc}
    \toprule
    {Task} & {Dataset} & {Composition} & {Query Size} & {Corpus Size} \\
    \midrule
    \multirow{10}{*}{I-CLS} 
     & ImageNet-1K~\cite{deng2009imagenet} & I2T & 1000 & 1000 \\
     & N24News~\cite{wang-etal-2022-n24news} & TI2T & 1000 & 24 \\
     & HatefulMemes~\cite{kiela2020hateful} & I2T & 1000 & 2 \\
     & VOC2007~\cite{everingham2015pascal} & I2T & 1000 & 20 \\
     & SUN397~\cite{xiao2010sun} & I2T & 1000 & 397 \\
     & Place365~\cite{zhou2017places} & I2T & 1000 & 365 \\
     & ImageNet-A~\cite{hendrycks2021natural} & I2T & 1000 & 1000 \\
     & ImageNet-R~\cite{hendrycks2021many} & I2T & 1000 & 200 \\
     & ObjectNet~\cite{barbu2019objectnet} & I2T & 1000 & 113 \\
     & Country211~\cite{radford2021learning} & I2T & 1000 & 211 \\
    \midrule
    \multirow{10}{*}{I-QA} 
     & OK-VQA~\cite{marino2019ok} & TI2T & 1000 & 1000 \\
     & A-OKVQA~\cite{schwenk2022okvqa} & TI2T & 1000 & 894 \\
     & DocVQA~\cite{mathew2021docvqa} & TI2T & 1000 & 1000 \\
     & InfographicsVQA~\cite{mathew2022infographicvqa} & TI2T & 1000 & 1000 \\
     & ChartQA~\cite{masry-etal-2022-chartqa} & TI2T & 1000 & 1000 \\
     & Visual7W~\cite{zhu2016visual7w} & TI2T & 1000 & 1000 \\
     & ScienceQA~\cite{lu2022learn} & TI2T & 1000 & 736 \\
     & VizWiz~\cite{gurari2018vizwiz} & TI2T & 1000 & 1000 \\
     & GQA~\cite{hudson2019gqa} & TI2T & 1000 & 1000 \\
     & TextVQA~\cite{singh2019towards} & TI2T & 1000 & 1000 \\
    \midrule
    \multirow{12}{*}{IT RET} 
     & MSCOCO\_i2t~\cite{lin2014microsoft} & I2T & 1000 & 1000 \\
     & VisualNews\_i2t~\cite{liu-etal-2021-visual} & I2T & 1000 & 1000 \\
     & VisDial~\cite{das2017visual} & T2I & 1000 & 1000 \\
     & MSCOCO\_t2i~\cite{lin2014microsoft} & T2I & 1000 & 1000 \\
     & VisualNews\_t2i~\cite{liu-etal-2021-visual} & T2I & 1000 & 1000 \\
     & WebQA~\cite{chang2022webqa} & T2TI & 1000 & 1000 \\
     & EDIS~\cite{liu-etal-2023-edis} & T2TI & 1000 & 1000 \\
     & Wiki-SS-NQ~\cite{ma-etal-2024-unifying} & T2I & 1000 & 1000 \\
     & CIRR~\cite{liu2021image} & TI2I & 1000 & 1000 \\
     & NIGHTS~\cite{fu2023dreamsim} & I2I & 1000 & 1000 \\
     & OVEN~\cite{hu2023open} & TI2TI & 1000 & 1000 \\
     & FashionIQ~\cite{wu2021fashion} & TI2I & 1000 & 1000 \\
    \midrule
    \multirow{4}{*}{VG} 
     & MSCOCO~\cite{lin2014microsoft} & TI2I & 1000 & 1000 \\
     & RefCOCO~\cite{kazemzadeh-etal-2014-referitgame} & TI2I & 1000 & 1000 \\
     & RefCOCO-Matching~\cite{kazemzadeh-etal-2014-referitgame} & TI2TI & 1000 & 1000 \\
     & Visual7W-Pointing~\cite{zhu2016visual7w} & TI2I & 1000 & 1000 \\
    \bottomrule
    \end{tabular}
}
\caption{Evaluation datasets for image modality, including task type and query-to-target composition.}
\label{tab:supple_eval_image}
\end{table*}

\begin{table*}[t]
\centering
\resizebox{.75\linewidth}{!}{%
\begin{tabular}{clccc}
\toprule
{Task} & {Dataset} & {Composition} & {Query Size} & {Corpus Size} \\
\midrule
\multirow{5}{*}{V-CLS} 
 & SmthSmthV2~\cite{goyal2017something} & V2T & 1000 & 174 \\
 & HMDB51~\cite{kuehne2011hmdb} & V2T & 1000 & 51 \\
 & UCF101~\cite{soomro2012ucf101} & V2T & 1000 & 101 \\
 & K700~\cite{carreira2019short} & V2T & 1000 & 700 \\
 & Breakfast~\cite{kuehne2014language} & V2T & 433 & 10 \\
\midrule
\multirow{5}{*}{T2V RET} 
 & MSR-VTT~\cite{xu2016msr} & T2V & 1000 & 1000 \\
 & MSVD~\cite{chen-dolan-2011-collecting} & T2V & 670 & 670 \\
 & DiDeMo~\cite{anne2017localizing} & T2V & 1004 & 1004 \\
 & YouCook2~\cite{zhou2018towards} & T2V & 3179 & 3179 \\
 & VATEX~\cite{wang2019vatex} & T2V & 4478 & 4478 \\
\midrule
\multirow{5}{*}{V2T RET} 
 & MSR-VTT~\cite{xu2016msr} & V2T & 1000 & 995 \\
 & MSVD~\cite{chen-dolan-2011-collecting} & V2T & 670 & 660 \\
 & DiDeMo~\cite{anne2017localizing} & V2T & 1004 & 1004 \\
 & YouCook2~\cite{zhou2018towards} & V2T & 3179 & 3105 \\
 & VATEX~\cite{wang2019vatex} & V2T & 4478 & 4478 \\
\midrule
\multirow{3}{*}{M-RET} 
 & QVHighlight~\cite{lei2021detecting} & TV2V & 1083 & 10 \\
 & Charades-STA~\cite{gao2017tall} & TV2V & 727 & 10 \\
 & MomentSeeker~\cite{NEURIPS2025_281e0b91} & TV2V & 1602 & 10 \\
\midrule
\multirow{5}{*}{V-VQA} 
 & Video-MME~\cite{Fu_2025_CVPR} & TV2T & 2700 & 4 \\
 & NExTQA~\cite{Xiao_2021_CVPR} & TV2T & 8564 & 5 \\
 & EgoSchema~\cite{NEURIPS2023_90ce332a} & TV2T & 500 & 5 \\
 & MVBench~\cite{Li_2024_CVPR} & TV2T & 4000 & 5 \\
 & ActivityNetQA~\cite{10.1609/aaai.v33i01.33019127} & TV2T & 1000 & 2 \\
\bottomrule
\end{tabular}
}
\caption{Evaluation datasets for video modality, including task type and query-to-target composition.}
\label{tab:supple_eval_video}
\end{table*}

\begin{table*}[t]
\centering
\resizebox{.75\linewidth}{!}{%
\begin{tabular}{clccc}
\toprule
{Task} & {Dataset} & {Composition} & {Query Size} & {Corpus Size} \\
\midrule
\multirow{3}{*}{A-CLS} 
 & ESC50~\cite{piczak2015esc} & A2T & 1237 & 50 \\
 & GTZAN~\cite{tzanetakis2002musical} & A2T & 290 & 10 \\
 & NSynth~\cite{engel2017neural} & A2T & 4096 & 10 \\
\midrule
\multirow{4}{*}{T2A RET} 
 & AudioCaps~\cite{kim-etal-2019-audiocaps} & T2A & 4411 & 883 \\
 & Clotho~\cite{bac89bfb28074b369efe4238b0048d5e} & T2A & 5225 & 1045 \\
 & MusicCaps~\cite{agostinelli2023musiclm} & T2A & 2772 & 2772 \\
 & SoundDescs~\cite{koepke2022audio} & T2A & 1000 & 1000 \\
\midrule
\multirow{3}{*}{A2T RET} 
 & AudioCaps~\cite{kim-etal-2019-audiocaps} & A2T & 883 & 4201 \\
 & Clotho~\cite{bac89bfb28074b369efe4238b0048d5e} & A2T & 1045 & 5225 \\
 & MusicCaps~\cite{agostinelli2023musiclm} & A2T & 2772 & 2772 \\
\midrule
\multirow{2}{*}{A-QA} 
 & MMAU~\cite{sakshi2025mmau} & TA2T & 1000 & 4 \\
 & MMAR~\cite{ma2026mmar} & TA2T & 995 & 4 \\
\bottomrule
\end{tabular}
}
\caption{Evaluation datasets for audio modality, including task type and query-to-target composition.}
\label{tab:supple_eval_audio}
\end{table*}

\begin{table*}[t]
\centering
\resizebox{.75\linewidth}{!}{%
\begin{tabular}{clccc}
\toprule
{Task} & {Dataset} & {Composition} & {Query Size} & {Corpus Size} \\
\midrule
\multirow{2}{*}{A2V RET} 
 & AVE~\cite{tian2018audio} & A2V & 402 & 402 \\
 & VALOR32k~\cite{10721284} & A2V & 3239 & 3239 \\
\midrule
\multirow{2}{*}{V2A RET} 
 & AVE~\cite{tian2018audio} & V2A & 402 & 402 \\
 & VALOR32k~\cite{10721284} & V2A & 3239 & 3239 \\
\midrule
\multirow{2}{*}{T2VA RET} 
 & AVHBench~\cite{sung-bin2025avhbench} & T2VA & 1105 & 1105 \\
 & VALOR32k~\cite{10721284} & T2VA & 3239 & 3239 \\
\midrule
\multirow{2}{*}{VA2T RET} 
 & AVHBench~\cite{sung-bin2025avhbench} & VA2T & 1105 & 1105 \\
 & VALOR32k~\cite{10721284} & VA2T & 3239 & 3239 \\
\midrule
\multirow{2}{*}{AV-QA} 
 & AVHBenchQA~\cite{sung-bin2025avhbench} & TVA2T & 5293 & 2 \\
 & DailyOmni~\cite{zhou2025daily} & TVA2T & 1196 & 4 \\
\bottomrule
\end{tabular}
}
\caption{Evaluation datasets for audiovisual modality, including task type and query-to-target composition.}
\label{tab:supple_eval_audiovisual}
\end{table*}

Intriguingly, the image-specific model exhibits solid performance on video tasks, while the video-specific model retains reasonable results on both image and audiovisual tasks. These trends indicate that a unified training setting, where a model is exposed to diverse modalities, can induce cross-modal synergy. This serves as the primary motivation behind the design of our Orthogonal Modality-Expert LoRA (OME-LoRA) paired with the Progressive Synergy Routing (PSR) mechanism.

While the \baseline\ baseline exhibits performance comparable to or slightly lower than that of the respective modality-specific models in their native modalities, our \mymethod\ framework consistently delivers improved performance, demonstrating the effectiveness of cross-modal synergy.

\subsection{Runtime Analysis}
To evaluate inference-time efficiency, we benchmark \baseline\ and \mymethod\ on the VALOR-32k audiovisual input using a single NVIDIA RTX A6000 GPU (batch size 1). As shown in~\cref{tab:runtime_analysis}, \mymethod\ introduces a slight increase in both latency and memory across 3B and 7B scales. Specifically, for the 3B variant, the mean latency increases from 300.3~ms to 354.6~ms, while the peak GPU memory changes from 9612~MB to 9676~MB.

This overhead is due to architectural differences at inference time. In \baseline, standard LoRA adapters are fully merged into the backbone weights, maintaining the original backbone's inference path. In contrast, while \mymethod\ merges its shared adaptation path, it explicitly retains the four modality-specific expert paths and router parameters to support input-dependent routing.

\subsection{Per-dataset Evaluation}
\label{sec:per_dataset}
We present the comprehensive benchmark-level results across all evaluated modalities and tasks. Individual dataset performances for image, video, audio, and audiovisual tasks are detailed in \cref{tab:indi_image}, \cref{tab:indi_video}, \cref{tab:indi_audio}, and \cref{tab:indi_audiovisual}, respectively.

\begin{table*}[t]
\centering
\resizebox{.90\linewidth}{!}{%
\begin{tabular}{lccc|cc|ccccc|cc}
\toprule
 & \rotatebox{90}{Nemotron (3B)} & \rotatebox{90}{LCO-Emb (3B)} & \rotatebox{90}{e5-omni (3B)} & \rotatebox{90}{\baseline\ (3B)} & \rotatebox{90}{\mymethod\ (3B)} & \rotatebox{90}{OmniEmbed (7B)} & \rotatebox{90}{Multivent (7B)} & \rotatebox{90}{LCO Emb (7B)} & \rotatebox{90}{e5-omni (7B)} & \rotatebox{90}{WAVE (7B)} & \rotatebox{90}{\baseline\ (7B)} & \rotatebox{90}{\mymethod\ (7B)} \\
\midrule
\textbf{Average} & & & & & & & & & & & & \\
Overall (36)         &                     44.1 &               58.1 &         64.8 &       66.8 &          68.1 &           44.2 &                     51.8 &               61.6 &         72.5 &      42.7 &             71.3 &          71.7 \\
I-CLS (10)  &   47.9 & 57.3 &    60.7 &  64.8 & 65.4 &  44.5 &   56.5 & 59.5 &    67.6 & 49.0 &    66.1 & 67.5 \\
I-QA (10)   &   20.1 & 58.2 &    61.3 &  63.1 & 62.7 &  22.5 &   29.6 & 62.7 &    69.9 & 25.8 &    67.8 & 66.8 \\
I-RET (12)  &   58.7 & 55.4 &    68.3 &  66.3 & 66.3 &  49.5 &   61.1 & 58.9 &    71.9 & 44.1 &    68.3 & 69.1 \\
VG  (4)  &   49.5 & 61.5 &    69.1 &  72.8 & 77.8 &  60.5 &   60.0 & 65.1 &    80.9 & 51.9 &    83.2 & 83.6 \\

\midrule
\textbf{Image Classification (I-CLS)} & & & & & & & & & & & & \\
ImageNet-1K     &   60.6 & 65.5 &    72.9 &  74.2 & 75.0 &  56.4 &   68.6 & 72.0 &    79.6 & 47.8 &    76.5 & 77.1 \\
N24News   &   36.9 & 50.0 &    73.1 &  73.4 & 77.6 &  42.5 &   42.3 & 45.6 &    77.9 & 43.3 &    76.5 & 80.2 \\
HatefulMemes   &   48.1 & 56.8 &    57.3 &  67.5 & 72.3 &  52.5 &   51.2 & 66.4 &    69.0 & 51.2 &    71.2 & 75.7 \\
VOC2007   &   53.0 & 66.4 &    71.8 &  87.5 & 88.2 &  75.1 &   68.3 & 54.4 &    83.3 & 77.2 &    87.9 & 89.3 \\
SUN397     &   54.8 & 73.8 &    71.2 &  78.0 & 76.1 &  46.1 &   61.6 & 73.1 &    75.0 & 62.7 &    76.6 & 77.2 \\
Place365  &   31.6 & 41.0 &    40.4 &  43.4 & 44.9 &  16.2 &   37.1 & 38.6 &    44.7 & 33.7 &    44.9 & 47.1 \\
ImageNet-A       &   41.6 & 45.4 &    43.3 &  53.4 & 51.8 &  29.9 &   52.4 & 53.0 &    60.4 & 24.5 &    57.6 & 58.0 \\
ImageNet-R       &   82.8 & 72.0 &    85.6 &  85.1 & 82.4 &  62.7 &   87.7 & 83.0 &    86.7 & 68.0 &    86.5 & 84.6 \\
ObjectNet &   46.2 & 73.2 &    65.5 &  55.5 & 55.7 &  37.8 &   62.9 & 75.4 &    68.2 & 60.9 &    52.3 & 53.3 \\
Country211       &   23.6 & 29.3 &    26.0 &  29.8 & 29.8 &  25.7 &   32.5 & 33.5 &    30.8 & 20.7 &    30.8 & 32.3 \\
\midrule
\textbf{Image QA (I-QA)} & & & & & & & & & & & & \\
OK-VQA &   17.4 & 57.8 &    62.7 &  65.2 & 66.8 &  24.2 &   29.3 & 62.0 &    70.4 & 32.7 &    70.5 & 70.4 \\
A-OKVQA     &   12.5 & 48.1 &    52.2 &  55.4 & 53.6 &  15.6 &   20.0 & 53.9 &    59.6 & 25.0 &    57.8 & 59.3 \\
DocVQA &   17.5 & 84.1 &    81.6 &  91.3 & 92.6 &  22.8 &   25.8 & 89.6 &    93.3 & 18.6 &    93.4 & 92.3 \\
InfographicsVQA &    8.5 & 58.6 &    49.9 &  68.2 & 63.5 &  11.7 &   12.4 & 66.6 &    68.9 & 17.0 &    74.7 & 72.9 \\
ChartQA     &   13.3 & 40.1 &    47.7 &  58.7 & 58.5 &  12.4 &   14.0 & 46.9 &    65.7 & 12.5 &    64.5 & 63.1 \\
Visual7W    &    7.4 & 48.5 &    55.2 &  49.5 & 48.1 &   4.8 &   11.9 & 47.9 &    63.7 & 18.7 &    53.3 & 50.8 \\
ScienceQA   &   25.3 & 44.4 &    48.7 &  43.2 & 43.9 &  25.5 &   33.2 & 52.2 &    56.9 & 21.8 &    54.4 & 52.6 \\
VizWiz &   34.8 & 54.6 &    52.5 &  52.3 & 51.0 &  26.0 &   41.8 & 54.1 &    55.9 & 31.2 &    54.8 & 52.1 \\
GQA     &   28.5 & 65.4 &    79.1 &  67.4 & 67.9 &  37.1 &   48.8 & 69.7 &    78.4 & 50.2 &    70.1 & 70.0 \\
TextVQA     &   36.1 & 80.6 &    83.0 &  79.5 & 81.2 &  44.9 &   58.5 & 84.3 &    85.9 & 29.9 &    84.0 & 84.0 \\
\midrule
\textbf{Image-Text Retrieval (I-RET)} & & & & & & & & & & & & \\
VisDial    &   52.8 & 50.5 &    77.8 &  71.0 & 73.9 &  50.6 &   56.2 & 54.7 &    80.8 & 19.7 &    74.8 & 76.1 \\
CIRR   &   13.2 & 42.9 &    41.1 &  47.8 & 48.8 &  14.4 &   16.2 & 47.1 &    49.3 & 31.4 &    48.0 & 52.5 \\
VisualNews\_t2i &   60.1 & 54.9 &    70.1 &  70.7 & 72.6 &  37.7 &   63.3 & 61.3 &    74.5 & 45.3 &    74.3 & 75.2 \\
VisualNews\_i2t &   58.4 & 59.4 &    76.4 &  73.4 & 75.1 &  43.8 &   64.2 & 61.1 &    82.3 & 36.3 &    75.7 & 77.3 \\
MSCOCO\_t2i       &   59.9 & 59.2 &    74.3 &  65.6 & 67.1 &  51.2 &   63.6 & 65.2 &    76.2 & 63.5 &    70.7 & 70.7 \\
MSCOCO\_i2t       &   56.2 & 59.7 &    71.7 &  67.2 & 67.4 &  40.7 &   59.0 & 62.0 &    74.1 & 55.1 &    70.1 & 69.7 \\
NIGHTS     &   64.6 & 56.7 &    66.3 &  68.5 & 67.3 &  59.2 &   63.7 & 61.3 &    64.8 & 58.2 &    67.2 & 67.6 \\
WebQA &   90.2 & 84.7 &    90.5 &  89.1 & 88.9 &  77.1 &   90.9 & 84.0 &    88.6 & 51.5 &    90.3 & 90.1 \\
FashionIQ &    8.6 & 14.9 &    16.5 &  19.9 & 19.1 &   8.3 &   11.4 & 19.3 &    19.4 &  6.5 &    18.8 & 22.9 \\
Wiki-SS-NQ       &   87.0 & 52.6 &    74.2 &  69.1 & 62.3 &  84.2 &   86.6 & 55.0 &    83.3 & 51.1 &    64.4 & 63.6 \\
OVEN   &   72.7 & 75.2 &    75.8 &  74.0 & 74.1 &  65.2 &   74.6 & 77.2 &    79.1 & 57.2 &    77.3 & 76.9 \\
EDIS   &   81.2 & 53.8 &    85.3 &  79.9 & 79.2 &  61.7 &   83.0 & 58.4 &    90.0 & 53.1 &    88.3 & 86.9 \\
\midrule
\textbf{Visual Grounding (VG)} & & & & & & & & & & & & \\
MSCOCO    &   35.1 & 45.5 &    50.9 &  67.9 & 68.0 &  40.0 &   42.4 & 44.2 &    67.2 & 25.1 &    74.1 & 76.1 \\
RefCOCO   &   55.1 & 68.8 &    76.6 &  75.3 & 84.8 &  67.8 &   65.9 & 73.5 &    83.7 & 49.7 &    87.3 & 91.0 \\
RefCOCO-Matching  &   65.3 & 75.7 &    79.2 &  80.3 & 85.6 &  74.6 &   69.8 & 82.6 &    91.6 & 83.1 &    91.6 & 90.6 \\
Visual7W-Pointing &   42.7 & 55.8 &    69.6 &  67.8 & 72.9 &  59.6 &   62.0 & 60.2 &    81.1 & 49.7 &    79.6 & 76.6 \\
\bottomrule
\end{tabular}
}
\caption{Individual performances on image modality tasks. The method order is identical to that in \cref{tab:method_comparison}.}
\label{tab:indi_image}
\end{table*}

\begin{table*}[t]
\centering
\resizebox{.93\linewidth}{!}{%
\begin{tabular}{lccc|cc|ccccc|cc}
\toprule
 & \rotatebox{90}{Nemotron (3B)} & \rotatebox{90}{LCO-Emb (3B)} & \rotatebox{90}{e5-omni (3B)} & \rotatebox{90}{\baseline\ (3B)} & \rotatebox{90}{\mymethod\ (3B)} & \rotatebox{90}{OmniEmbed (7B)} & \rotatebox{90}{Multivent (7B)} & \rotatebox{90}{LCO Emb (7B)} & \rotatebox{90}{e5-omni (7B)} & \rotatebox{90}{WAVE (7B)} & \rotatebox{90}{\baseline\ (7B)} & \rotatebox{90}{\mymethod\ (7B)} \\
\midrule
\textbf{Average} & & & & & & & & & & & & \\
Overall (23)         &   36.5 &   43.7 &   40.6 &   40.4 &   40.5 &   35.0 &   40.2 &   45.5 &   44.3 &   39.8 &   40.2 &   41.4 \\
V-CLS (5)        &   43.1 &   44.6 &   37.6 &   47.6 &   44.7 &   36.4 &   50.7 &   47.6 &   49.4 &   48.9 &   49.5 &   47.0 \\
T2V RET (5)      &   34.8 &   34.2 &   38.9 &   27.9 &   29.3 &   33.7 &   35.9 &   36.9 &   36.1 &   30.0 &   25.0 &   28.5 \\
V2T RET (5)      &   32.0 &   34.3 &   33.2 &   38.3 &   39.2 &   27.9 &   34.7 &   35.9 &   40.8 &   35.1 &   40.5 &   41.6 \\
M-RET (3)        &   25.6 &   47.2 &   41.0 &   36.9 &   37.5 &   28.7 &   28.1 &   47.1 &   34.1 &   40.1 &   32.4 &   36.1 \\
V-QA (5)         &   46.9 &   58.1 &   52.0 &   51.1 &   51.7 &   48.3 &   51.8 &   60.2 &   61.1 &   44.9 &   53.5 &   53.8 \\
\midrule
\textbf{Video Classification (V-CLS)} & & & & & & & & & & & & \\
K700               &   38.5 &   42.6 &   42.4 &   46.3 &   43.9 &   31.5 &   53.3 &   43.2 &   50.1 &   51.4 &   52.0 &   48.0 \\
SmthSmthV2       &   40.0 &   44.1 &   35.2 &   45.9 &   45.9 &   34.6 &   54.8 &   46.5 &   46.1 &   47.2 &   48.8 &   51.1 \\
HMDB51           &   48.0 &   53.2 &   34.1 &   52.1 &   47.3 &   26.2 &   53.3 &   57.7 &   56.3 &   53.4 &   54.7 &   44.2 \\
UCF101           &   62.7 &   64.2 &   51.9 &   71.5 &   63.1 &   62.2 &   68.2 &   67.8 &   69.5 &   72.0 &   66.6 &   68.3 \\
Breakfast       &   26.3 &   18.7 &   24.5 &   22.2 &   23.1 &   27.3 &   24.0 &   22.6 &   24.9 &   20.3 &   25.4 &   23.6 \\
\midrule
\textbf{Text $\to$ Video Retrieval (T2V RET)} & & & & & & & & & & & & \\
DiDeMo        &   36.0 &   33.1 &   41.6 &   27.4 &   27.4 &   27.6 &   34.7 &   34.7 &   35.7 &   25.6 &   23.1 &   27.3 \\
MSR-VTT       &   37.8 &   37.4 &   40.0 &   39.6 &   43.1 &   40.2 &   40.5 &   41.2 &   42.1 &   34.6 &   37.6 &   43.9 \\
MSVD          &   55.8 &   58.4 &   62.1 &   43.3 &   46.7 &   56.4 &   56.9 &   60.0 &   56.7 &   50.9 &   39.9 &   45.1 \\
VATEX         &   28.8 &   29.1 &   34.0 &   25.5 &   24.3 &   28.1 &   30.1 &   32.3 &   30.8 &   29.4 &   23.0 &   21.3 \\
YouCook2     &   15.5 &   13.0 &   16.8 &    3.7 &    5.0 &   16.0 &   17.4 &   16.1 &   15.3 &    9.7 &    1.7 &    4.7 \\
\midrule
\textbf{Video $\to$ Text Retrieval (V2T RET)} & & & & & & & & & & & & \\
DiDeMo    &   32.0 &   32.4 &   31.9 &   42.5 &   42.0 &   27.7 &   35.7 &   33.6 &   43.0 &   32.7 &   44.3 &   46.4 \\
MSR-VTT  &   34.3 &   38.8 &   37.6 &   42.4 &   45.8 &   33.2 &   37.3 &   37.9 &   44.2 &   39.0 &   47.0 &   48.1 \\
MSVD      &   53.6 &   56.9 &   57.6 &   60.7 &   61.5 &   49.7 &   56.9 &   60.1 &   63.1 &   58.4 &   62.5 &   63.7 \\
VATEX     &   25.6 &   28.3 &   28.4 &   29.9 &   31.4 &   21.5 &   28.5 &   30.0 &   37.2 &   30.8 &   31.6 &   32.0 \\
YouCook2  &   14.4 &   15.4 &   10.6 &   15.8 &   15.3 &    7.4 &   15.1 &   17.6 &   16.4 &   14.7 &   17.3 &   18.0 \\
\midrule
\textbf{Moment Retrieval (M-RET)} & & & & & & & & & & & & \\
QVHighlight     &   26.9 &   70.3 &   60.4 &   51.9 &   53.4 &   29.9 &   30.0 &   69.2 &   44.1 &   56.0 &   41.5 &   50.9 \\
Charades-STA    &   11.6 &   25.2 &   19.5 &   18.0 &   16.6 &   14.3 &   13.1 &   24.3 &   16.9 &   26.4 &   16.8 &   14.9 \\
MomentSeeker    &   38.3 &   46.1 &   43.1 &   40.8 &   42.6 &   41.9 &   41.2 &   47.8 &   41.4 &   38.0 &   39.0 &   42.7 \\
\midrule
\textbf{Video QA (V-QA)} & & & & & & & & & & & & \\
MVBench          &   42.3 &   58.0 &   51.5 &   50.9 &   53.1 &   43.6 &   47.2 &   60.9 &   60.1 &   39.3 &   54.7 &   54.5 \\
Video-MME        &   40.1 &   46.1 &   43.0 &   41.3 &   41.6 &   38.4 &   40.9 &   49.7 &   49.0 &   32.1 &   43.6 &   44.0 \\
NExTQA           &   43.9 &   67.4 &   58.9 &   55.5 &   59.4 &   45.9 &   48.1 &   70.3 &   71.1 &   38.4 &   60.8 &   60.0 \\
EgoSchema        &   46.6 &   43.8 &   45.6 &   34.8 &   30.6 &   46.4 &   51.4 &   42.8 &   51.4 &   48.0 &   34.2 &   35.4 \\
ActivityNetQA    &   61.6 &   75.1 &   61.1 &   72.9 &   73.9 &   67.3 &   71.5 &   77.3 &   74.1 &   66.5 &   74.2 &   75.1 \\
\bottomrule
\end{tabular}
}
\caption{Individual performances on video modality tasks. The method order is identical to that in \cref{tab:method_comparison}.}
\label{tab:indi_video}

\end{table*}

\begin{table*}[t]
\centering
\resizebox{.910\linewidth}{!}{%
\begin{tabular}{lccc|cc|ccccc|cc}
\toprule
 & \rotatebox{90}{Nemotron (3B)} & \rotatebox{90}{LCO-Emb (3B)} & \rotatebox{90}{e5-omni (3B)} & \rotatebox{90}{\baseline\ (3B)} & \rotatebox{90}{\mymethod\ (3B)} & \rotatebox{90}{OmniEmbed (7B)} & \rotatebox{90}{Multivent (7B)} & \rotatebox{90}{LCO Emb (7B)} & \rotatebox{90}{e5-omni (7B)} & \rotatebox{90}{WAVE (7B)} & \rotatebox{90}{\baseline\ (7B)} & \rotatebox{90}{\mymethod\ (7B)} \\
\midrule
\textbf{Average} & & & & & & & & & & & & \\
Overall (12)          &   24.5 &   42.1 &   37.4 &   43.1 &   45.6 &   31.6 &   39.5 &   45.2 &   44.3 &   33.4 &   47.5 &   48.7 \\
A-CLS (3)         &   40.7 &   67.5 &   51.5 &   59.5 &   62.0 &   36.0 &   52.9 &   71.1 &   62.8 &   56.3 &   67.4 &   67.2 \\
T2A RET (4)       &    6.4 &   19.7 &   26.0 &   28.7 &   30.8 &   25.5 &   28.6 &   24.5 &   30.2 &   22.5 &   32.5 &   34.9 \\
A2T RET (3)       &    6.8 &   17.3 &   14.2 &   30.0 &   32.4 &   17.2 &   21.0 &   18.7 &   20.7 &   17.7 &   32.5 &   34.0 \\
A-QA (2)          &   44.0 &   63.8 &   57.9 &   54.4 &   57.0 &   47.8 &   55.4 &   66.4 &   63.7 &   37.0 &   57.6 &   58.5 \\
\midrule
\textbf{Audio Classification (A-CLS)} & & & & & & & & & & & & \\
ESC50                 &   73.6 &   77.8 &   78.2 &   86.8 &   87.0 &   51.5 &   77.8 &   78.8 &   81.9 &   82.8 &   88.0 &   88.5 \\
GTZAN                 &   28.6 &   74.8 &   44.1 &   52.1 &   57.9 &   34.5 &   51.4 &   82.8 &   61.4 &   57.9 &   66.9 &   63.4 \\
NSynth                &   20.0 &   50.0 &   32.1 &   39.6 &   41.0 &   22.0 &   29.5 &   51.6 &   45.0 &   28.1 &   47.3 &   49.7 \\
\midrule
\textbf{Text $\to$ Audio Retrieval (T2A RET)} & & & & & & & & & & & & \\
AudioCaps             &    9.4 &   23.3 &   33.9 &   39.9 &   43.2 &   39.0 &   33.5 &   26.9 &   38.6 &   32.1 &   41.2 &   44.9 \\
Clotho                &    7.2 &   17.2 &   21.6 &   20.4 &   20.6 &   21.8 &   21.5 &   19.8 &   22.7 &   19.1 &   21.3 &   21.8 \\
MusicCaps             &    3.5 &    8.0 &   10.5 &   19.8 &   23.4 &   11.8 &   13.2 &   11.9 &   14.8 &    8.7 &   20.6 &   22.6 \\
SoundDescs            &    5.3 &   30.4 &   38.1 &   34.6 &   36.1 &   29.4 &   46.2 &   39.3 &   44.5 &   29.9 &   46.9 &   50.5 \\
\midrule
\textbf{Audio $\to$ Text Retrieval (A2T RET)} & & & & & & & & & & & & \\
AudioCaps        &   10.6 &   26.0 &   26.0 &   49.4 &   52.8 &   33.7 &   34.8 &   30.0 &   36.8 &   26.6 &   54.2 &   57.6 \\
Clotho            &    7.4 &   17.4 &   13.3 &   23.4 &   23.6 &   15.2 &   19.4 &   18.0 &   17.1 &   19.1 &   26.9 &   26.5 \\
MusicCaps        &    2.3 &    8.3 &    3.3 &   17.2 &   20.7 &    2.7 &    8.9 &    8.2 &    8.0 &    7.3 &   16.4 &   18.0 \\
\midrule
\textbf{Audio QA (A-QA)} & & & & & & & & & & & & \\
MMAU                  &   50.7 &   70.5 &   63.9 &   59.9 &   62.1 &   50.0 &   58.7 &   74.9 &   70.5 &   36.1 &   64.8 &   61.8 \\
MMAR                  &   37.4 &   57.1 &   52.0 &   48.8 &   52.0 &   45.5 &   52.1 &   58.0 &   56.9 &   38.0 &   50.5 &   55.3 \\
\bottomrule
\end{tabular}
}
\caption{Individual performances on audio modality tasks. The method order is identical to that in \cref{tab:method_comparison}.}
\label{tab:indi_audio}

\end{table*}

\begin{table*}[t]
\centering
\resizebox{.910\linewidth}{!}{%
\begin{tabular}{lccc|cc|ccccc|cc}
\toprule
 & \rotatebox{90}{Nemotron (3B)} & \rotatebox{90}{LCO-Emb (3B)} & \rotatebox{90}{e5-omni (3B)} & \rotatebox{90}{\baseline\ (3B)} & \rotatebox{90}{\mymethod\ (3B)} & \rotatebox{90}{OmniEmbed (7B)} & \rotatebox{90}{Multivent (7B)} & \rotatebox{90}{LCO Emb (7B)} & \rotatebox{90}{e5-omni (7B)} & \rotatebox{90}{WAVE (7B)} & \rotatebox{90}{\baseline\ (7B)} & \rotatebox{90}{\mymethod\ (7B)} \\
\midrule
\textbf{Average} & & & & & & & & & & & & \\
Overall (10)          &   28.5 &   35.5 &   31.6 &   40.4 &   43.4 &   29.0 &   34.4 &   35.5 &   42.3 &   38.1 &   43.9 &   44.6 \\
A2V RET (2)       &    5.2 &   11.1 &    7.0 &   17.4 &   19.1 &    5.2 &   11.4 &   12.6 &    7.2 &    8.8 &   22.0 &   23.1 \\
V2A RET (2)       &    7.3 &   10.5 &    9.2 &   14.3 &   17.9 &   13.5 &   16.1 &   13.8 &   14.8 &   19.9 &   20.1 &   18.9 \\
T2VA RET (2)      &   46.9 &   51.1 &   64.4 &   63.3 &   64.2 &   57.2 &   48.9 &   51.7 &   65.7 &   52.6 &   64.7 &   64.9 \\
VA2T RET (2)      &   41.4 &   46.6 &   30.5 &   62.5 &   63.0 &   27.3 &   49.1 &   46.3 &   62.4 &   56.9 &   64.3 &   63.9 \\
AV-QA (2)         &   41.5 &   58.4 &   46.9 &   44.8 &   52.6 &   41.8 &   46.7 &   53.3 &   61.6 &   52.4 &   48.5 &   52.0 \\
\midrule
\textbf{Audio $\to$ Video Retrieval (A2V RET)} & & & & & & & & & & & & \\
AVE              &    6.7 &   12.9 &   10.2 &   18.2 &   19.4 &    8.2 &   11.2 &   15.2 &    9.5 &   10.0 &   23.4 &   23.9 \\
VALOR32k         &    3.7 &    9.2 &    3.9 &   16.6 &   18.8 &    2.2 &   11.6 &    9.9 &    5.0 &    7.7 &   20.6 &   22.2 \\
\midrule
\textbf{Video $\to$ Audio Retrieval (V2A RET)} & & & & & & & & & & & & \\
AVE              &   10.9 &   12.7 &   10.9 &   12.9 &   19.9 &   16.4 &   17.4 &   14.7 &   15.9 &   22.1 &   20.6 &   15.9 \\
VALOR32k         &    3.7 &    8.4 &    7.5 &   15.6 &   16.0 &   10.5 &   14.8 &   13.0 &   13.6 &   17.7 &   19.6 &   21.9 \\
\midrule
\textbf{Text $\to$ Audiovisual Retrieval (T2VA RET)} & & & & & & & & & & & & \\
AVHBench      &   56.2 &   60.6 &   75.0 &   74.9 &   75.7 &   67.3 &   57.6 &   61.1 &   76.5 &   62.4 &   75.4 &   76.0 \\
VALOR32k       &   37.6 &   41.6 &   53.8 &   51.6 &   52.7 &   47.1 &   40.1 &   42.4 &   54.9 &   42.7 &   54.0 &   53.9 \\
\midrule
\textbf{Audiovisual $\to$ Text Retrieval (VA2T RET)} & & & & & & & & & & & & \\
AVHBench       &   50.2 &   53.2 &   23.6 &   73.5 &   74.8 &   36.6 &   58.6 &   54.6 &   71.1 &   67.1 &   74.8 &   75.4 \\
VALOR32k        &   32.6 &   39.9 &   37.4 &   51.6 &   51.3 &   18.1 &   39.6 &   38.0 &   53.8 &   46.8 &   53.8 &   52.4 \\
\midrule
\textbf{Audiovisual QA (AV-QA)} & & & & & & & & & & & & \\
AVHBenchQA             &   53.4 &   69.8 &   55.8 &   63.8 &   65.5 &   52.2 &   60.7 &   60.5 &   79.5 &   76.2 &   60.2 &   66.7 \\
DailyOmni              &   29.5 &   47.0 &   38.1 &   25.8 &   39.7 &   31.4 &   32.6 &   46.2 &   43.8 &   28.6 &   36.8 &   37.2 \\
\bottomrule
\end{tabular}
}
\caption{Individual performances on audiovisual modality tasks. The method order is identical to that in \cref{tab:method_comparison}.}
\label{tab:indi_audiovisual}

\end{table*}

\section{AI Tool Usage}
We utilized ChatGPT to improve the writing fluency and grammar of the manuscript. The development of the core ideas, technical methodologies, experimental designs, and analyses remains the original work of the authors.

\section{Artifact}
We meticulously document the licenses, access URLs, and intended uses of the artifacts utilized in this study. All datasets, models, and software artifacts were used in accordance with their original licenses and intended research usage conditions when specified. For redistributed repositories or mirrors, we additionally verified the original artifact licenses when available. All derivatives or model checkpoints in our work are targeted for non-commercial research purposes.

{\setdefaultleftmargin{1mm}{}{}{}{}{}
\begin{itemize}

\item \textbf{MMEB v1 Training Set}~\cite{jiang2025vlmvec}: Sourced from the official HuggingFace repository\footnote{\url{https://huggingface.co/datasets/TIGER-Lab/MMEB-train}} under the \texttt{Apache-2.0} license.

\item \textbf{LLaVAHound}~\cite{zhang-etal-2025-direct}: Sourced from the official HuggingFace repository\footnote{\url{https://huggingface.co/datasets/ShareGPTVideo/train_video_and_instruction}} under the \texttt{Apache-2.0} license.

\item \textbf{PE-Video}~\cite{bolya2026perception}: Sourced from the official HuggingFace repository\footnote{\url{https://huggingface.co/datasets/facebook/PE-Video}} under the \texttt{CC-BY-NC-4.0} license for non-commercial academic research.

\item \textbf{MSVD}~\cite{chen-dolan-2011-collecting} \& \textbf{MSR-VTT}~\cite{xu2016msr}: Sourced from a public redistributed HuggingFace repository\footnote{\url{https://huggingface.co/datasets/friedrichor/MSVD}}\footnote{\url{https://huggingface.co/datasets/friedrichor/MSR-VTT}} while original licenses unspecified.

\item \textbf{ActivityNet Captions}~\cite{caba2015activitynet,krishna2017dense}: Sourced from a redistributed HuggingFace repository\footnote{\url{https://huggingface.co/datasets/friedrichor/ActivityNet_Captions}} under the \texttt{Apache License 2.0}.

\item \textbf{DiDeMo}~\cite{anne2017localizing}: Sourced from a redistributed HuggingFace repository\footnote{\url{https://huggingface.co/datasets/friedrichor/DiDeMo}} under the \texttt{BSD 2-Clause} license.

\item \textbf{FineCVR}~\cite{wu2025learning}: Sourced from the official GitHub repository\footnote{\url{https://github.com/May2333/FDCA}}, while license information is unknown.

\item \textbf{AudioCaps}~\cite{kim-etal-2019-audiocaps}: Sourced from a redistributed HuggingFace mirror\footnote{\url{https://huggingface.co/datasets/CLAPv2/audiocaps}} under the \texttt{MIT} license.

\item \textbf{WavCaps}~\cite{mei2024wavcaps}: Sourced from the official HuggingFace repository\footnote{\url{https://huggingface.co/datasets/cvssp/WavCaps}} under the \texttt{CC-BY-4.0} license.

\item \textbf{AudioSet Strong}~\cite{gemmeke2017audio}: Sourced from a redistributed HuggingFace mirror\footnote{\url{https://huggingface.co/datasets/CLAPv2/audioset_strong}} under the \texttt{CC BY 4.0} specification.

\item \textbf{MusicCaps}~\cite{agostinelli2023musiclm}: Sourced from a redistributed HuggingFace mirror\footnote{\url{https://huggingface.co/datasets/CLAPv2/MusicCaps}} under the \texttt{CC-BY-SA-4.0} framework.

\item \textbf{MusicBench}~\cite{melechovsky-etal-2024-mustango}: Sourced from the official HuggingFace repository\footnote{\url{https://huggingface.co/datasets/amaai-lab/MusicBench}} under the \texttt{CC-BY-SA-3.0} license.

\item \textbf{MusicInstruct}~\cite{deng-etal-2024-musilingo}: Sourced from the official HuggingFace repository\footnote{\url{https://huggingface.co/datasets/m-a-p/Music-Instruct}} under the \texttt{CC-BY-NC-4.0} license.

\item \textbf{AudioSet}~\cite{gemmeke2017audio}: Sourced from the official page\footnote{\url{https://research.google.com/audioset/}} under the \texttt{CC BY 4.0} license.

\item \textbf{VAST}~\cite{chen2023vast}: Sourced from the official GitHub repository\footnote{\url{https://github.com/CASIA-IVA-Lab/VAST}} under the \texttt{MIT} license.

\item \textbf{InternVideo2}~\cite{wang2024internvideo2}: Sourced from the official HuggingFace repository\footnote{\url{https://huggingface.co/datasets/OpenGVLab/InternVideo2_Vid_Text}} under the \texttt{CC-BY-NC-SA-4.0} license.

\item \textbf{Shot2Story}~\cite{han2025shot2story}: Sourced from the official HuggingFace repository\footnote{\url{https://huggingface.co/ByteDance/shot2story}} under the \texttt{CC-BY-NC-SA-4.0} license, restricted to non-commercial academic usage.

\item \textbf{AVInstruct}~\cite{ye2024cat}: Sourced from the official GitHub repository\footnote{\url{https://github.com/rikeilong/Bay-CAT}} under the \texttt{Apache-2.0} license, utilizing video sources derived from AVQA~\cite{yang2022avqa}.

\item \textbf{MMEB v1 Evaluation Set}~\cite{jiang2025vlmvec}: Sourced from the official HuggingFace repository\footnote{\url{https://huggingface.co/datasets/TIGER-Lab/MMEB-eval}} under the \texttt{Apache-2.0} license.

\item \textbf{MMEB v2 Evaluation Set}~\cite{meng2026vlmvecv}: Sourced from the official HuggingFace repository\footnote{\url{https://huggingface.co/datasets/TIGER-Lab/MMEB-V2}} under the \texttt{Apache-2.0} protocol.

\item \textbf{ESC-50}~\cite{piczak2015esc}: Sourced from a redistributed HuggingFace mirror\footnote{\url{https://huggingface.co/datasets/CLAPv2/esc50_no_overlap}} under the \texttt{CC-BY-NC} license.

\item \textbf{GTZAN}~\cite{tzanetakis2002musical}: Sourced from a redistributed HuggingFace mirror\footnote{\url{https://huggingface.co/datasets/m-a-p/GTZAN}}.

\item \textbf{NSynth}~\cite{engel2017neural}: Sourced from a redistributed HuggingFace mirror\footnote{\url{https://huggingface.co/datasets/TwinkStart/Nsynth}} under the \texttt{CC BY 4.0} license.

\item \textbf{AudioCaps (for Eval)}~\cite{kim-etal-2019-audiocaps}: Sourced from a redistributed HuggingFace mirror\footnote{\url{https://huggingface.co/datasets/OpenSound/AudioCaps}} under the \texttt{MIT} license.

\item \textbf{Clotho}~\cite{bac89bfb28074b369efe4238b0048d5e}: Sourced from the official archive\footnote{\url{https://zenodo.org/records/4783391}} under its custom academic non-commercial research license.

\item \textbf{SoundDescs}~\cite{koepke2022audio} \& \textbf{AVE}~\cite{tian2018audio}: Sourced from a redistributed HuggingFace repository\footnote{\url{https://huggingface.co/datasets/VLM2Vec/MMEB-V3}}. \textbf{SoundDescs} follows the \texttt{Apache License 2.0}, while the license information for \textbf{AVE} was not explicitly provided in the original repository.

\item \textbf{MMAU}~\cite{sakshi2025mmau}: Sourced from a redistributed HuggingFace repository\footnote{\url{https://huggingface.co/datasets/gamma-lab-umd/MMAU-test-mini}} under the \texttt{CC-BY-SA-4.0} standards.

\item \textbf{MMAR}~\cite{ma2026mmar}: Sourced from the official HuggingFace repository\footnote{\url{https://huggingface.co/datasets/BoJack/MMAR}} under the \texttt{CC-BY-NC-4.0} license.

\item \textbf{VALOR 32k}~\cite{10721284}: Sourced from the official GitHub repository\footnote{\url{https://github.com/CASIA-IVA-Lab/VALOR}} under the \texttt{MIT} framework.

\item \textbf{AVHBench}~\cite{sung-bin2025avhbench}: Sourced from the official GitHub repository\footnote{\url{https://github.com/kaist-ami/AVHBench}}.

\item \textbf{DailyOmni}~\cite{zhou2025daily}: Sourced from a redistributed HuggingFace repository\footnote{\url{https://huggingface.co/datasets/mteb/Daily-Omni}} under the \texttt{GPL-3.0} license.

\item \textbf{Qwen 2.5 Omni}~\cite{xu2025qwen25}: Sourced from the official HuggingFace repository\footnote{\url{https://huggingface.co/Qwen/Qwen2.5-Omni-7B}} under the \texttt{Apache-2.0} license.

\item \textbf{Omni-Embed-Nemotron}~\cite{xu2025omni}: Sourced from the official HuggingFace repository\footnote{\url{https://huggingface.co/nvidia/omni-embed-nemotron-3b}} under NVIDIA's custom non-commercial scientific license (\texttt{customized-nscl-v1}) for non-commercial research and scientific evaluation.

\item \textbf{LCO-Emb}~\cite{xiao2026scaling}: Sourced from the official HuggingFace repository\footnote{\url{https://huggingface.co/LCO-Embedding}} under the \texttt{Apache-2.0} policy.

\item \textbf{e5-omni}~\cite{chen-etal-2026-e5}: Sourced from the official HuggingFace repository\footnote{\url{https://huggingface.co/collections/Haon-Chen/e5-omni}} under the \texttt{MIT} license.

\item \textbf{OmniEmbed-v0.1} \& \textbf{OmniEmbed-Multivent}~\cite{ma2025tevatron, zhan2025magmar}: Sourced from the official HuggingFace repository\footnote{\url{https://huggingface.co/Tevatron/OmniEmbed-v0.1}}
\footnote{\url{https://huggingface.co/Tevatron/OmniEmbed-v0.1-multivent}} under the \texttt{MIT} license.

\item \textbf{WAVE}~\cite{tang2026wave}: Sourced from the official GitHub repository\footnote{\url{https://github.com/TCL606/WAVE}} under the \texttt{Apache License 2.0}.

\item \textbf{GradCache}~\cite{gao-etal-2021-scaling}: Sourced from the official GitHub repository\footnote{\url{https://github.com/luyug/GradCache}} under the \texttt{Apache-2.0} license.

\item \textbf{Flash-Attention-2}~\cite{dao2023flashattention}: Sourced from the official GitHub repository\footnote{\url{https://github.com/dao-ailab/flash-attention}} under the \texttt{BSD-3-Clause} license.

\item \textbf{qwen-omni-utils}: Sourced from the official GitHub repository\footnote{\url{https://github.com/qwenlm/qwen2.5-omni}} under the \texttt{Apache-2.0} license.

\item \textbf{Tevatron 2.0}~\cite{ma2025tevatron}: Sourced from the official GitHub repository\footnote{\url{https://github.com/texttron/tevatron}} under the \texttt{Apache-2.0} protocol.
\end{itemize}
}

All datasets were obtained from public benchmarks and repositories, and we relied on the original providers' release procedures for potentially identifiable or offensive content.

\end{document}